\pdfoutput=1

\documentclass[11pt]{article}

\usepackage[preprint]{acl}

\usepackage[T1]{fontenc}
\usepackage[utf8]{inputenc}
\usepackage{times}
\usepackage{latexsym}

\usepackage{microtype}
\usepackage{inconsolata}
\usepackage{graphicx}

\usepackage{booktabs}
\usepackage{multirow}
\usepackage{trajan}

\usepackage{xspace}
\usepackage{amssymb}
\usepackage{amsmath}
\usepackage{svg}
\usepackage{subfigure}
\usepackage{xcolor}
\usepackage{float}
\usepackage{tcolorbox}

\usepackage{graphicx}
\usepackage{subcaption}
\usepackage{tikz}
\usepackage{mdframed}
\usepackage{tcolorbox}
\tcbuselibrary{breakable}
\usepackage{xcolor}
\usepackage{lipsum}

\usepackage{afterpage}
\usepackage{enumitem}
\usepackage{cuted}
\usepackage[normalem]{ulem}

\usepackage{supertabular}
\usepackage{array}
\usepackage{etoc}

\definecolor{mygray}{HTML}{333333}
\definecolor{myblue}{HTML}{0048B5}
\definecolor{mygreen}{HTML}{608C28}
\definecolor{myyellow}{HTML}{BF9000}

\newcommand{\model}{{\trjnfamily SOTOPIA}\xspace}
\newcommand{\modelhard}{{\trjnfamily SOTOPIA}-{hard}\xspace}
\newcommand{\mymodel}{{\trjnfamily SOTOPIA}-{$\Omega$}\xspace}
\newcommand{\sotopiapi}{{\trjnfamily SOTOPIA}-{$\pi$}\xspace}
\newcommand{\eval}{ {\trjnfamily SOTOPIA-EVAL}\xspace}
\newcommand{\mistral}{{{Mistral-7B}}\xspace}

\renewcommand{\scriptsize}{\fontsize{8pt}{8pt}\selectfont}

\title{ \mymodel: Dynamic Strategy Injection Learning and Social Instruction Following Evaluation for Social Agents}

\author{
    Wenyuan Zhang$^{\P,\S}$, Tianyun Liu$^{\P}$, Mengxiao Song$^{\P,\S}$, 
    \textbf{Xiaodong Li}$^{\P,\S}$, \textbf{Tingwen Liu}$^{\P,\S,}$\thanks{~indicates corresponding author.} \\
  $^{\P}$Institute of Information Engineering, Chinese Academy of Sciences, China \\
  $^{\S}$School of Cyber Security, University of Chinese Academy of Sciences, China \\
  \texttt{\{zhangwenyuan,liutianyun,songmengxiao,lixiaodong,liutingwen\}@iie.ac.cn}\\\\
  \href{https://github.com/WYRipple/SOTOPIA-Omega}
  {\textbf{Code}}\quad \href{https://huggingface.co/datasets/WYRipple/sotopia-omega}{\textbf{Data}}\quad\href{https://huggingface.co/collections/WYRipple/sotopia-checkpoints-683527f070d215849adc2873}{\textbf{Checkpoints}}\\
}

\begin{document}
\maketitle
\begin{abstract}
Despite the abundance of prior social strategies humans possess, there remains a paucity of research dedicated to their transfer and integration into social agents.
Our proposed \mymodel framework aims to address and bridge this gap, with a particular focus on enhancing the social capabilities of language agents. 
This framework dynamically injects multi-step reasoning strategies inspired by negotiation theory and two simple direct strategies into expert agents, thereby automating the construction of a high-quality social dialogue training corpus. 
Additionally, we introduce the concept of Social Instruction Following (S-IF) and propose two new S-IF evaluation metrics that complement social capability. 
We demonstrate that several 7B models trained on high-quality corpus significantly surpass the expert agent (GPT-4) in achieving social goals and enhancing S-IF performance.
Analysis and variant experiments validate the advantages of dynamic construction, which can especially break the agent's prolonged deadlock.

\end{abstract}

\section{Introduction}

Recently, studies on the social simulation of large language model intelligent agents have been growing interest~\cite{richards2023principlist,park2023generative,choi-etal-2023-llms,wang2024demo}.
By assigning identities~\cite{chen-etal-2024-socialbench} and social goals~\cite{zhang2024imperative}, intelligent agents are anticipated to exhibit advanced human-like social abilities~\cite{huang2023humanity}, such as emotional care~\cite{van2023emotion}, collaboration~\cite{lan-etal-2024-llm} and negotiation~\cite{abdelnabi2024llmdeliberation}.

\begin{figure}[!t]
\centering  
\includegraphics[width=7.5cm]{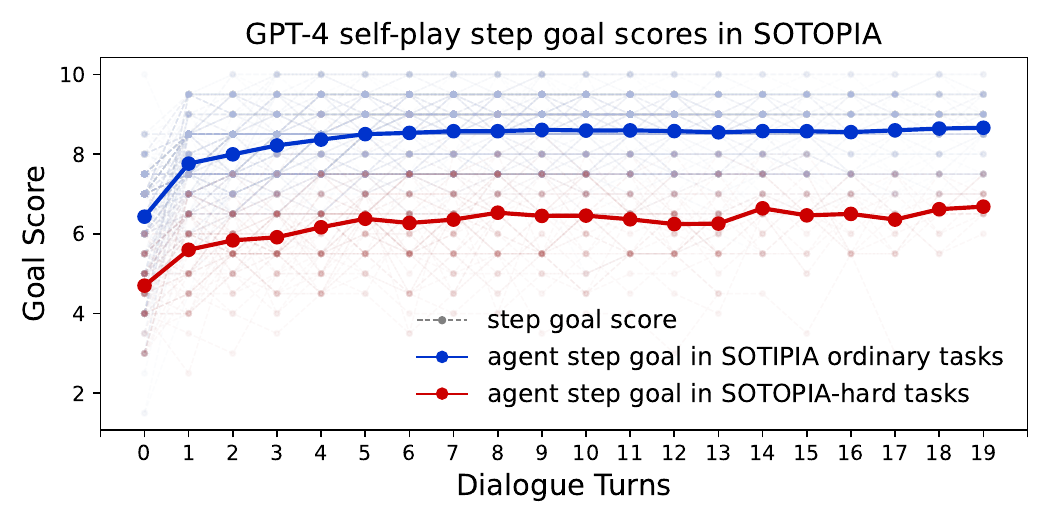}
\caption{The average \textsc{Goal} scores per turn in GPT-4 self-play are shown for 70 hard and 380 ordinary tasks in \model~\cite{zhou2024sotopia}. Goal measures how well each agent achieves its social goal during interaction. In both settings, expert agents struggle to significantly improve their goal scores after only a few of turns. More details are provided in Sec~\S\ref{sec:sotopia_intro}.}
\label{fig:motivation}
\end{figure}

However, existing research~\cite{zhou2024sotopia} shows that even expert agents\footnote{\citet{zhou2024sotopia} validates GPT-4's strong social capability and refers to it as an expert agent.} perform significantly worse on challenging social tasks compared to ordinary tasks. 
Moreover, as shown in Figure~\ref{fig:motivation}, our turn-level evaluation of expert self-play reveals that goal scores remain nearly unchanged after only a few turns. 
This phenomenon has been described as prolonged deadlocks by \citet{narlikar2010deadlocks}.

The main reason for these phenomena is the conflict between the social goals of the two agents, as they remain fixated on their current viewpoints and struggle to identify potential win-win strategies~\cite{thompson2015mind}.
A natural approach is to inject prior strategies into the social agent\footnote{In this paper, we collectively refer to the LLMs involved in the final social task inference as social agents.}. 
However, many existing methods struggle to adapt to open-ended social tasks, either constraining the action space~\cite{deng2023plug,zhang-etal-2024-strength} or being tailored to specific tasks~\cite{feng2023towards,chang-chen-2024-injecting}.
\citet{sotopia-pi} introduces a promising framework for training social agents, leveraging both innate expert strategies and self-generated ones through behavior cloning and self-reinforcement.
However, it also inherits behaviors that cause prolonged deadlocks, ultimately capping the social agent's performance at the expert's original level.

To overcome the limitations of existing approaches, we propose \mymodel, a dynamic strategy injection framework for generating high-quality social dialogue corpus.
To overcome the issue of prolonged deadlocks in the expert agent, we design a negotiation strategy injection workflow inspired by negotiation theory~\cite{thompson2015mind}. 
Inspired by the principles of slow-thinking~\cite{min2024imitate,lin2024swiftsage}, the workflow adopts a structured, multi-step reasoning approach to assist experts in identifying potential win-win strategies in scenarios involving conflicting goals.
Meanwhile, we retain the expert's native strategies or apply simple strategy guidance to prevent over-reasoning~\cite{chiang2024over}. 
Additionally, the framework introduces step rating as a self-supervised reward~\cite{yangrewards}, ensuring dynamic strategy selection and adjustment during dialogue generation.

Furthermore, we introduce Social Instruction Following (S-IF), which is the capability of social agents to follow instructions in goal-driven tasks. 
Preliminary experiments show that even expert agents exhibit ``parroting'' (repeating actions) and ``topic drift'' (generating off-goal content), revealing limitations in fundamental generative capabilities and their disconnect from goal achievement. 
To address this, we propose two turn-level metrics: $S_{div}$, penalizing overly similar actions, and $S_{rel}$, measuring action relevance to the goal.

Experiments show that social agents trained with Dynamic Strategy Injection (DSI-learning) outperform expert agents like GPT-4 in social capabilities.
DSI-learning also boosts S-IF capabilities, leading to more diverse outputs and goal-aligned actions. 
The generated corpus demonstrates significantly higher goal scores, reducing deadlock issues.
Variant experiments confirm DSI-learning's superiority over non-dynamic settings, with negligible impact on generalization and safety.

Our contributions can be summarized as follows:
\begin{itemize}
    \setlength{\itemsep}{0pt}
    \setlength{\parskip}{0pt}
    \setlength{\parsep}{0pt}
    \item We propose a novel dynamic strategy injection framework, \mymodel, for generating social dialogue corpora.  
    \item We introduce the concept of Social Instruction Following and propose two new metrics.
    \item Extensive experiments validate the superiority of \mymodel. 
    \item We will open-source the high-quality dialogue corpus with step ratings and robust social agent weights to the community.
\end{itemize}

\section{Related Work}
We provide a comprehensive literature review in Appendix~\S\ref{sub_app:related_work}, while this section focuses primarily on the most relevant works.

\subsection{Social Agent and Strategy Injection}
Large language models (LLMs) can potentially become proficient social agents~\cite{park2023generative,huang2023humanity,choi-etal-2023-llms}. 
However, accurately simulating diverse and open-ended human social behaviors in an infinite action space remains a challenge ~\cite{mou2024individual}.
As a result, task-specific agents~\cite{chen2023money} or those with constrained actions~\cite{deng2023plug,zhang-etal-2024-strength} struggle to adapt.

Strategy injection enhances social agents by integrating human priors into model behavior. 
Existing work mainly focuses on \textit{inference-time injection}, such as multi-agent interactions~\cite{lan-etal-2024-llm} or auxiliary strategy models~\cite{chang-chen-2024-injecting,feng2023towards}, but these methods incur significant inference overhead and are task-specific~\cite{deng2023plug}. 
In contrast, \citet{sotopia-pi} introduces \textit{training-time injection}, cloning expert behavior and applying self-reinforcement. 
This allows weaker agents to achieve expert-level performance in open-ended, multi-task settings without additional inference costs.
However, it inherits the expert’s limitations, potentially leading to issues like prolonged deadlock.

\subsection{Negotiation Theory}
\label{sec:related_work_negotiation}

Negotiation theory~\cite{korobkin2024negotiation} offers a universal strategy framework for addressing social tasks, many of which are characterized as mixed-motive negotiations~\cite{deutsch1973resolution}. 
These involve non-adversarial interactions where parties have differing motivations and preferences~\cite{froman1970compromise}. 
Even when social goals seem to conflict, a win-win outcome can be achieved by finding complementary interests.
According to negotiation theory, final agreements in such scenarios can approach the Pareto frontier~\cite{tripp1992evaluation}, though achieving such outcomes remains challenging. 
To address this, \citet{thompson2015mind} proposes a structured workflow that guides negotiators toward Pareto-optimal solutions. 
Based on this framework, we distill four core steps applicable to social tasks: \textit{Resource Assessment}, \textit{Assessment of Difference}, \textit{Initial Proposal} and \textit{Update Proposal}.

\section{Preliminaries}
\label{sec:sotopia_intro}
\model~\cite{zhou2024sotopia} is an open environment designed to test the social ability of social agents, which includes 450 diverse social tasks, with a challenging subset of 70 designated as \modelhard.
Specifically, each task is assigned a social profile that includes two social agents' personal details with their goals and a scenario.
In each task, two agents take turns acting. 
The first acting agent is denoted as $\pi_1$ and the second as $\pi_2$.
Each action constitutes a turn $i$, abbreviated as $a^{\pi}_{i}$. 
Thus, the interaction output can be formalized as a sequence $\{a^{\pi_1}_{0},a^{\pi_2}_{1},a^{\pi_1}_{2},...\}$, 
and $a^{\pi_{[1,2]}}_{i}\sim \pi_{[1,2]}^{\theta_{[1,2]} } (\cdot |p^{\pi_{[1,2]}}\oplus a_{<i}^{\pi})$ represents the action of the agent parameterized by $\theta_{i}$.
Where {\small $[\cdot,\cdot]$} represents alternating agent index, $p$ denotes agent's profile, $a_{<i}^{\pi}$ represents the history of actions before turn $i$, and $\oplus$ indicates concatenation.
The ultimate goal is to obtain an agent with a better $\theta$.
Appendix~\S\ref{sub_app:sotopia_details} details the environment.

\section{\mymodel Framework}
\label{sec:workflow}
\begin{figure*}[!h]  
\centering  
\includegraphics[width=16cm]{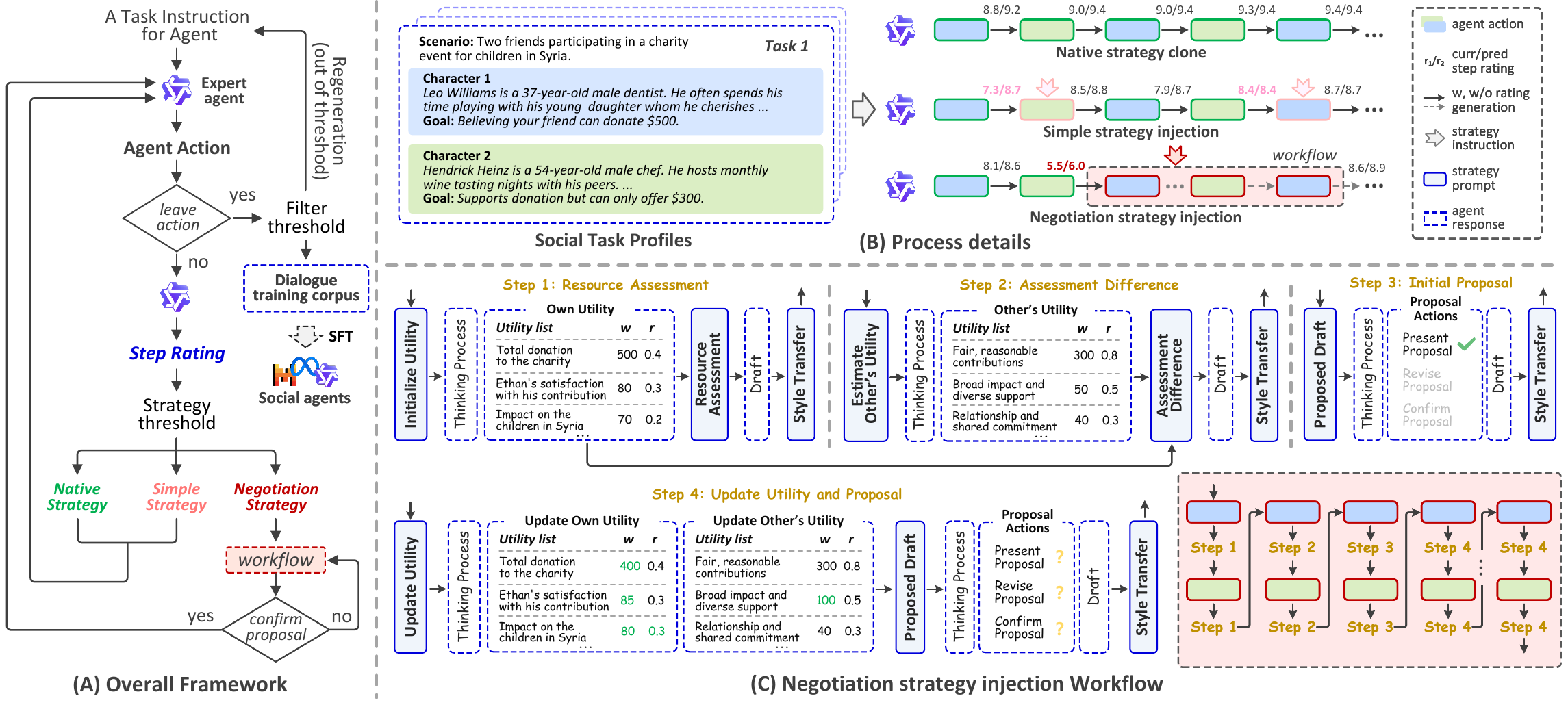}
\caption{The architecture and details of \mymodel. (A) represents the overall architecture for data generation. (B) provides the step rating details of (A), demonstrating the process of injecting three strategies. (C) illustrates the negotiation strategy injection workflow, where the input at each step is the current dialogue history, and the output is the final response. The bottom-right corner shows the input-output flow of the negotiation strategy.}
\label{fig:main}
\end{figure*}

\mymodel is a dialogue data generation framework designed to dynamically inject advanced strategies into the data generation process, enabling the creation of a high-quality training corpus.
As illustrated in Figure~\ref{fig:main}, our framework incorporates two key mechanisms to help agents break through prolonged deadlocks: (1) three types of strategy injection methods for data generation and (2) dynamic strategy selection guided by step ratings.
Unlike inference-time injection systems~\cite{min2024imitate}, our framework is a behavior cloning approach~\cite{bain1995framework} that combines fast and slow thinking for data generation.
The data generation model, referred to as the expert agent, is powered by Qwen2.5-72B\footnote{https://huggingface.co/Qwen/Qwen2.5-72B-Instruct}, an open-source model selected for its robust social reasoning capabilities~\cite{yang2024qwen2,mou2024agentsense}.

\subsection{Strategy Injection Generation}
\mymodel employs three strategies, including two fast-thinking strategies and one slow-thinking workflow, to address prolonged deadlocks caused by varying degrees of conflict.

\subsubsection{Native Strategy Clone}
The expert agent exhibits effective strategies for ordinary social tasks~\cite{ijcai2024p890}, such as norm situations~\cite{ziems2023normbank}.
In this scenario, we retain the expert's native action to avoid introducing unnecessarily complex inference, akin to fast-thinking~\cite{de2023advancing}.

\subsubsection{Simple Strategy Injection}
Perspective-taking is an essential strategy in negotiation theory that guides both parties in a dialogue toward potential win-win outcomes~\cite{trotschel2011perspective}.
In tasks with minor conflicts, perspective-taking prompts foster altruistic behavior~\cite{underwood1982perspective}, effectively alleviating the endowment effects~\cite{galin2009proposal}.
This simple strategy, which does not introduce extra inference steps, can also be considered fast thinking\footnote{Simply adding an extra prompt: \textit{In your response, you should pay additional attention to the potential conflicts between you and the other party and work towards improving them, so that both sides can achieve their goals. Implicitly express the ``conflict'' while employing advanced linguistic techniques to propose ways of improvement, ensuring a smooth connection with the previous flow of conversation.}}.

\subsubsection{Negotiation Strategy Injection}
The expert agent exhibits significant limitations in resolving intense conflicts.
We leverage negotiation theory to develop the \textit{Negotiation strategy injection workflow}\footnote{We provide a case and explanation in Appendix~\S\ref{appendix:case} to help understand the negotiation strategy injection workflow.}, which allows the two parties in dynamic games reach a potential win-win outcome.
Specifically, as illustrated in Figure~\ref{fig:main}(C), the workflow comprises four steps, each requiring multiple turns of expert reasoning, making it a characteristic slow-thinking inference workflow~\cite{lin2024swiftsage}.
All generation includes a prior thought process, which has been shown to effectively improve quality~\cite{wei2022chain,wu2024thinking}.
In addition, we adopt a ``\textit{draft}-then-\textit{style-transfer}'' approach to enhance diversity while ensuring that responses adhere to predefined requirements.

\paragraph{Step 1: Resource Assessment}

The expert first presents its interests, formalized as a utility function, which is widely used in economics and game theory~\cite{houthakker1950revealed,slantchev2012game}:
\begin{equation}
    U=\frac{1}{n} {\sum_{i=1}^{n}}w_i r_i u_i,
\end{equation}
where $u_i$ represents the utility item, while $w_i$ and $r_i$ represent the weights and ratios. 
The utility function is expert-self-defined, and introducing more parameters can expand the expert's decision space, enhancing its versatility.
The utility is stored in JSON format and is further converted into a natural language description as the final response.

\paragraph{Step 2: Assessment of Difference}
In this step, the expert first guesses the opponent's utility based on their response. 
It then combines its own utility to jointly assess the potential conflicts in social goals between both parties, emphasizing items  of high value to itself but low value to the opponent.

\paragraph{Step 3: Initial Proposal}
Based on previous analysis, the agent presents its initial proposal and encourages identifying win-win strategies.
Both parties need to present their initial proposals, considering their differing positions.

\paragraph{Step 4: Update Proposal}
Both parties often struggle to reach a consensus on initial proposals that have not been discussed. 
To address this, an update mechanism dynamically adjusts utilities based on the opponent's proposal, aiming to identify a nearly complementary distribution of interests more accurately.
The agent then has three options.
\textit{Present Proposal}: If the agent deems the opponent’s proposal unlikely to yield a win-win outcome, it offers a new perspective.
\textit{Revise Proposal}: If the agent agrees with the opponent’s perspective but identifies details requiring adjustment, it refines the proposal.
\textit{Confirm Proposal}: If the agent considers the opponent’s proposal a viable win-win solution, it accepts the proposal.
Unlike the restricted actions~\cite{he2018decoupling,stasaski2020cima}, these options provide general response directions, encouraging the expert agent to explore available strategies independently.
After confirming the proposal, both parties exit the workflow and leave the conversation following uninterrupted interaction.

\subsection{Dynamic Strategy Selection}
This section introduces the core concepts of dynamic strategy selection, with all design details available in Appendix~\S\ref{sub_app:threshold}.

\subsubsection{Step Rating}
Step rating serves as a self-supervised reward \cite{yangrewards}, representing the expert's self-evaluation after generating each $a_i^{\pi}$.
The step rating encompasses the current goal and predicted goal, represented as $goal_c$ and $goal_p$.
$goal_c$ emphasizes the extent to which the current goal is achieved, while $goal_p$ introduces future expectations as a complementary measure. 
Higher values of both $goal_c$ and $goal_p$ can serve as evidence of the potential to achieve a future win-win outcome.

\begin{figure*}[!ht]
    \centering
    \includegraphics[width=\textwidth]{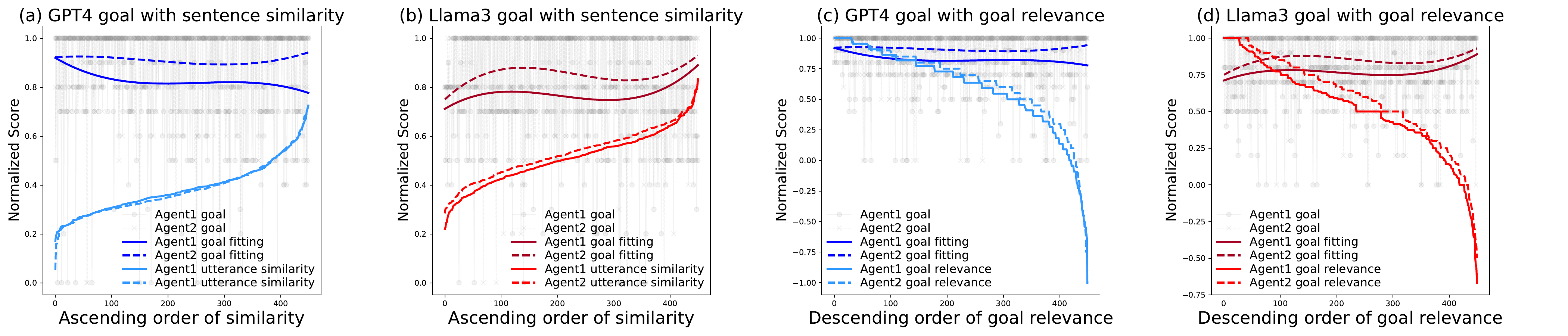}%
    \vspace{0.3em}
    \includegraphics[width=\textwidth]{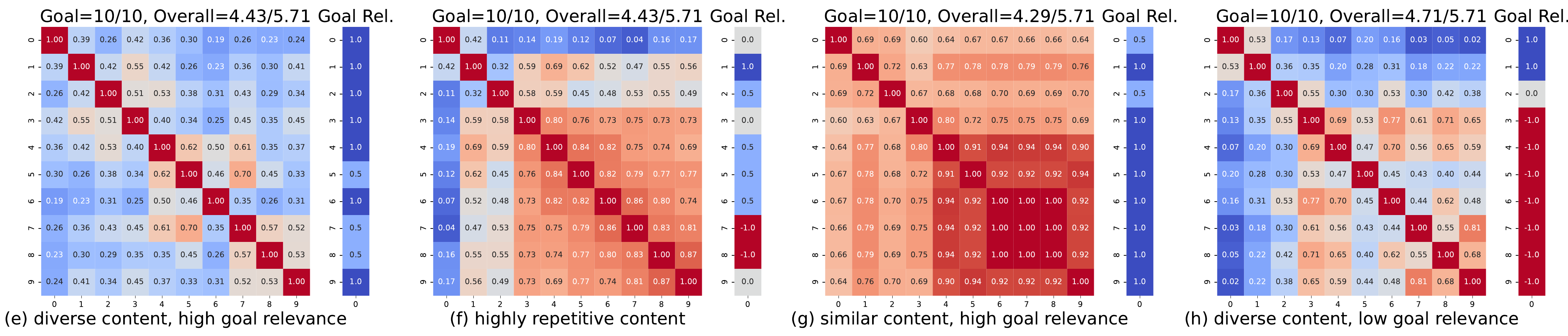}
    \caption{The pre-experiment for social instruction following evaluation uses Llama3-8B and GPT-4. (a-d) illustrate the relationship between action diversity, topic relevance, and goal scores across 450 tasks in \model, with goal curves fitted using a third-order polynomial. (e-h) present four cases from Llama3-8B. In the heatmaps, the left side shows the cosine similarity matrix of all actions one agent generates in a single task. In contrast, the right side indicates the goal relevance of each action, with red denoting poor performance.}
    \label{fig:two_long_pdfs}
\end{figure*}

\subsubsection{Dynamic Strategy Injection}



As shown in Figure~\ref{fig:main}, we implement dynamic strategy selection based on step rating. 
The expert performs six self-play turns in all training tasks to initialize the interaction (as illustrated in Figure~\ref{fig:motivation}, deadlocks typically occur after the sixth turn). 
Starting from the sixth action, the expert generates a step rating for each action and the interaction history, and determines the strategy for the next action based on three threshold intervals defined by the strategy threshold function. 
\textit{Native} and \textit{Simple Strategy} require re-evaluation at each turn, but once \textit{Negotiation Strategy} is selected, the corresponding workflow is activated, and step rating stops until the workflow completes. 
When the maximum number of interaction turns is reached or a ``leave'' action is generated, each complete dialogue undergoes a filtering function to assess quality.
Low-quality dialogues are flagged for regeneration. 
After all tasks are completed, a final high-quality social dialogue corpus is obtained.
The strategy and filter threshold function are detailed in Appendix~\ref{sub_app:threshold}.

Using the high-quality dialogue training corpus obtained from \mymodel, we directly fine-tune various social agents, and refer to them as dynamic strategy injection learning (DSI-learning) agents.
This enriches their internal strategies and enhances their social capabilities during dialogue inference without requiring any external intervention.

\begin{table*}[!t]
\scriptsize
    \centering
    \begin{tabular*}{0.96\textwidth}{@{\extracolsep{\fill}}@{}l cccc cccc @{}}
    {}  & \multicolumn{4}{c}{\textbf{\textit{GPT3.5 as Reference Model}}} & \multicolumn{4}{c}{\textbf{\textit{Self-play}}} \\
    \toprule
    \multirow{2}{*}{\raisebox{-0.5\height}{Agent model}} & \multicolumn{2}{c}{\textbf{\model}} & \multicolumn{2}{c}{\modelhard} & \multicolumn{2}{c}{\textbf{\model}} & \multicolumn{2}{c}{\modelhard} \\
    \cmidrule(lr){2-3}\cmidrule(lr){4-5}\cmidrule(lr){6-7}\cmidrule(lr){8-9}
    & \textsc{Goal $\uparrow$} & Overall $\uparrow$ & \textsc{Goal $\uparrow$} & Overall $\uparrow$ & \textsc{Goal $\uparrow$} & Overall $\uparrow$ & \textsc{Goal $\uparrow$} & Overall $\uparrow$ \\
    \midrule
    GPT-4~\cite{zhou2024sotopia}$^{\dagger \ddagger}$ & \underline{7.62} & 3.31 & \underline{5.92} & \underline{2.79} & \underline{8.60} & \underline{3.91} & 6.66 & 3.29\\
    GPT-3.5~\cite{zhou2024sotopia}$^{\dagger \ddagger}$ & 6.45 & 2.93 & 4.39 & 2.29 & 6.77 & 2.82 & 5.11 & 2.22 \\
    \midrule
    Base (\mistral)$^{\dagger \ddagger}$ & 5.07 & 2.33 & 3.84 & 1.98 & 4.27 & 1.62 & 3.41 & 1.33 \\
    BC+SR~\cite{sotopia-pi}$^{\dagger \ddagger}$ & \underline{7.62} & \underline{3.44} & 5.34 & 2.76 & 8.25 & 3.81 & \underline{7.07} & \underline{3.53} \\
    \midrule
    Dynamic Strategy Injection (DSI)  & \textbf{8.07} & \textbf{3.67} & \textbf{6.31} & \textbf{3.03} & \textbf{8.73} & \textbf{4.11} & \textbf{7.28} & \textbf{3.65} \\
    \% Improve (relative) &5.91\% & 6.69\% & 18.16\% & 9.78\% & 5.82\% & 7.87\% & 2.97\% & 3.40\% \\
    \bottomrule
    \end{tabular*}
    \caption{DSI enables a weak 7B social agent to surpass expert-level performance. The symbol $\dagger$ represents the performance reported in the original paper, while $\ddagger$ indicates the performance obtained through reproduction. Bold indicates the best, and underline indicates the second best. Our report results are the averages of the three evaluation outcomes (statistically significant with p < 0.05).}
    \label{tab:main_results}
\end{table*}

\section{Social Instruction Following Evaluation}
In this section, we introduce the concept of \underline{S}ocial \underline{I}nstruction \underline{F}ollowing (S-IF) for LLMs and propose two evaluation metrics.
Unlike traditional single/multi-turn instruction following, S-IF is characterized by three features: (1) multi-agent and multi-turn dialogue simulation, (2) agents with social identities, and (3) agents with social goals.
An agent with strong S-IF capabilities can generate diverse interactions that closely align with its social identity and goals while being able to exit the interaction appropriately.
This is distinct from general social capabilities (e.g., as measured by~\eval, see Sec~\S\ref{sec:experiment}). Our preliminary experiments have revealed this phenomenon.

Specifically, the preliminary experiments focus on two aspects: \textit{generative diversity} and \textit{goal relevance}. 
Regarding \textit{diversity}, one agent sometimes exhibits a ``parroting'' phenomenon in social tasks, characterized by high similarity among multiple actions within its own dialogue (Figure~\ref{fig:two_long_pdfs}(a,b)).
Even expert agents display this behavior, yet their social performance does not degrade with reduced diversity and may even improve unexpectedly. 
Regarding \textit{goal relevance}, agents frequently experience ``topic drift'', generating content mainly unrelated to the intended goal (Figure~\ref{fig:two_long_pdfs}(c,d)). 
Despite this off-topic behavior, their ability to achieve social goals remains unaffected.

Several cases illustrate the causes of the orthogonality phenomenon in Figure~\ref{fig:two_long_pdfs}(e-h).
(e) serves as an exemplary case of high social capability and strong S-IF performance. 
In contrast, (f) exhibits partially similar yet goal-irrelevant behaviors yet achieves win-win outcomes through limited goal-relevant actions. 
Meanwhile, (g) and (h) demonstrate the orthogonality of the two S-IF aspects. 
Specifically, even when agents achieve perfect goal scores, their actions may exhibit high similarity with repetitive, goal-relevant content (akin to ``parroting'') or deviate entirely after only two rounds of interaction despite generating diverse content.

We formally propose two metrics for evaluating S-IF ability: action diversity $S_{div}$ and goal relevance $S_{rel}$.
$S_{div}$ penalizes actions with extremely high similarity and scales the differences in high similarity scores through an external power function. 
A recommended $\alpha$ value is 10, which penalizes dialogues with only excessively similar actions\footnote{Appendix~\ref{app:action_sim} presents the impact of different $\alpha$ values on the results.}.
$S_{rel}$ averages the GPT-4 evaluation scores $goal(\cdot)$ of a set of dialogues and applies sigmoid normalization $\sigma (\cdot)$:
\begin{align}\label{eq:softpool1}
\small
\begin{split}
    & S_{div}=(\frac{1}{n-1}\sum_{i\ne j}^{n}[1-sim(a^\pi_{i},a^\pi_{j})^{\alpha}])^{\alpha}, \\
    & S_{rel}=\sigma( { \sum_{i=1}^{n}}goal(a^\pi_{i}) / n), \\
    & S_{sif}=\frac{1}{2} (S_{div}+S_{rel}),
\end{split}
\end{align}
where $S_{sif}$ serves as a comprehensive measure of the social agent's S-IF capability.

As an essential supplement, we provide a detailed discussion in Appendix~\S\ref{appendix:SIF}, covering the motivation for proposing S-IF and related work. We also analyze the differences between our proposed metrics and previous evaluation methods (e.g., ROSCOE~\cite{golovnevaroscoe}) and present all details of the preliminary experiments and the two metrics $S_{div}$, $S_{rel}$.

\section{Experiment Settings}
\label{sec:experiment}

\noindent\textbf{Training}
To ensure fairness, we use the open-source social task profiles\footnote{https://huggingface.co/datasets/cmu-lti/sotopia-pi}, constructed from \sotopiapi, as inputs for our framework.
These tasks are orthogonal to the tasks in the \model environment and include 410 scenarios.
The final generated corpus is used to fine-tune smaller social agents, referred to as DSI-learning agents.
Each agent is fine-tuned using LoRA on a single NVIDIA A100-80G, with training accelerated by Unsloth.
More details refer to Appendix~\ref{app:trining_settings}.

\noindent\textbf{Baseline}
BC+SR~\cite{sotopia-pi} is a two-stage training strategy that enables the social agent to learn from GPT-4 and its own highly rated behaviors.
BC+SR has open-sourced the fine-tuned base agent (Mistral-7B), which we use as a baseline. 
We also evaluate the social capabilities of Llama3-8B and Qwen2.5-7B as additional baselines.

\begin{table}[!t]
\setlength\tabcolsep{4.5pt}
\scriptsize
    \centering
    \begin{tabular}{l cccc}
    \toprule
    \multirow{2}{*}{\raisebox{-0.5\height}{Agent model}} & \multicolumn{2}{c}{\textbf{\model}} & \multicolumn{2}{c}{\modelhard} \\
    \cmidrule(lr){2-3}\cmidrule(lr){4-5}
    & \textsc{Goal} & Overall & \textsc{Goal } & Overall \\
    \midrule
    GPT-4 & 8.60 & 3.91 & 6.66 & 3.29  \\
    \midrule
    Llama3-8B  & 8.12 & 3.74 & 6.44 &  3.06  \\
    DSI  & \textbf{8.63} & \textbf{4.04} & \textbf{7.34} & \textbf{3.64}  \\
    \midrule
    Qwen2.5-7B  & 8.45 & 3.90 & 6.52 & 3.22  \\
    DSI  & \textbf{8.91} & \textbf{4.18} & \textbf{7.86} & \textbf{3.97}  \\
    
    \bottomrule
    \end{tabular}
    \caption{The performance of powerful social agents in the self-play setting. Bold indicates surpassing GPT-4.}
    \label{tab:more_model_results}
\end{table}

\begin{table*}[!t] 
\setlength{\tabcolsep}{5pt}
\scriptsize 
    \centering 
    \begin{tabular}{l ccc ccc ccc  c} 
        \toprule 
        
        \multirow{2}{*}{\raisebox{-0.6\height}{\parbox{0.8cm}{ Agent\\model}}} &
        
        \multicolumn{3}{c}{agent1} & \multicolumn{3}{c}{agent2} &  
        \multirow{2}{*}{\raisebox{-0.5\height}{$S_{div}$ Avg.$\uparrow$}} &
        \multirow{2}{*}{\raisebox{-0.5\height}{$S_{rel}$ Avg.$\uparrow$}} &
        \multirow{2}{*}{\raisebox{-0.5\height}{$S_{sif}$ Avg.$\uparrow$}} & 
        \multirow{2}{*}{\raisebox{-0.5\height}{\textit{Turns} Avg.$\downarrow$}} \\
        
        \cmidrule(lr){2-4}\cmidrule(lr){5-7}
        & $S_{div}\uparrow$ & $S_{rel}\uparrow$ & $S_{sif}\uparrow$ & $S_{div}\uparrow$ & $S_{rel}\uparrow$ & $S_{sif}\uparrow$ & & & & \\
        
        \midrule 
        GPT-4 & 91.60$^*$ & 62.77 & 77.19$^*$ & 91.33$^*$ & 64.43 & 77.88$^*$ & 91.47$^*$ & 63.60 & 77.54$^*$ & 15.2$^*$ \\ 
        \midrule
        Base & 15.69 & 57.11 & 36.40 & 18.82 & 54.27 & 36.55 & 17.25 & 55.69 & 36.48  & 19.7 \\ 
        BC+SR & 76.60 & 55.64 & 66.12 & 78.22 & 57.17 & 67.70 & 77.41 & 56.41 & 66.91 & 16.7 \\ 
        DSI & \textbf{79.90} & \textbf{63.94} & \textbf{71.92} & \textbf{79.50} & \textbf{65.26} & \textbf{72.38} & 
        \textbf{79.70}$_{\textcolor{blue}{\scalebox{0.6}{$\uparrow$} 62.45}}$ & 
        \textbf{64.60}$_{\textcolor{blue}{\scalebox{0.6}{$\uparrow$} 8.91}}$ &
        \textbf{72.15}$_{\textcolor{blue}{\scalebox{0.6}{$\uparrow$} 35.67}}$ & 
        \textbf{15.8}$_{\textcolor{blue}{\scalebox{0.6}{$\downarrow$} 3.9}}$ \\ 
        \midrule 
        Llama3-8B & 70.12 & 62.21 & 66.17 & 68.03 & 63.79 & 65.91 & 69.08 & 63.00 & 66.04 &  19.9 \\ 
        DSI & \textbf{79.25} & \textbf{64.47}$^*$ & \textbf{71.86} & \textbf{78.78} & \textbf{65.51}$^*$ & \textbf{72.15} & 
        \textbf{79.01}$_{\textcolor{blue}{\scalebox{0.6}{$\uparrow$} 9.93}}$ & 
        \textbf{64.99}$_{\textcolor{blue}{\scalebox{0.6}{$\uparrow$} 1.99}}^*$ & 
        \textbf{72.00}$_{\textcolor{blue}{\scalebox{0.6}{$\uparrow$} 5.96}}$ & 
        \textbf{17.6}$_{\textcolor{blue}{\scalebox{0.6}{$\downarrow$} 2.3}}$\\ 
        \midrule 
        Qwen2.5-7B & 81.22 & 60.99 & 71.11 & 79.52 & 62.29 & 70.91 & 80.37 & 61.64 & 71.01 & 20.0 \\ 
        DSI & \textbf{82.36} & \textbf{63.49} & \textbf{72.93} & \textbf{81.39} & \textbf{64.82} & \textbf{73.11} & 
        \textbf{81.87}$_{\textcolor{blue}{\scalebox{0.6}{$\uparrow$} 1.50}}$ & 
        \textbf{64.15}$_{\textcolor{blue}{\scalebox{0.6}{$\uparrow$} 2.51}}$ & 
        \textbf{73.02}$_{\textcolor{blue}{\scalebox{0.6}{$\uparrow$} 2.01}}$ &
        \textbf{17.0}$_{\textcolor{blue}{\scalebox{0.6}{$\downarrow$} 3.0}}$ \\ 
        
        \bottomrule
    \end{tabular}
    \caption{Social Instruction Following (S-IF) capability. Subscript blue numbers indicate the absolute improvement over the original social agent's performance. Bold indicates the best performance within an agent, while an asterisk denotes the best performance across all comparisons.} 
\label{tab:S_IF_results} 
\end{table*}

\noindent\textbf{Evaluation}
\eval~\cite{zhou2024sotopia} provides seven evaluation dimensions for social capabilities, in which \textsc{Goal} score represents the extent to which the agent achieves the social goal (an integer from 0 to 10).
We report average \textsc{Goal} and overall score (the average of seven dimensions) from 450 tasks in \model and provide the full results in the Appendix~\S\ref{appendix:detail_results}.
\textit{GPT3.5 as the reference model} refers to the interaction between the social agent and the weaker reference model GPT-3.5, reporting the performance of the social agent. 
\textit{Self-play} refers to the social agent interacting with itself, reporting the average results to explore the upper bound of its capability.
We follow sotopia-related research~\cite{sotopia-pi,zhou-etal-2024-real} and employ GPT-4 as the evaluator, as it has been validated to correlate highly with human evaluations~\cite{zhou2024sotopia}.
For more detailed experimental settings, refer to Appendix~\S\ref{appendix:setting_details}.

\begin{table*}[!t]
\setlength{\tabcolsep}{3.5pt}
\scriptsize
    \centering
    \begin{tabular*}{1\textwidth}{@{\extracolsep{\fill}}@{}l l cccc cccc @{}}
    {} & {} & \multicolumn{4}{c}{\textbf{\textit{GPT3.5 as Reference Model}}} & \multicolumn{4}{c}{\textbf{\textit{Self-play}}} \\
    \toprule
    \multirow{2}{*}{\raisebox{-0.7\height}{\parbox{1.7cm}{ Construction\\strategy}}} &
    \multirow{2}{*}{\raisebox{-0.7\height}{\parbox{1.7cm}{ \#Data volume\\(usable ratio)}}} &
    \multicolumn{2}{c}{\textbf{\model}} & \multicolumn{2}{c}{\modelhard} & \multicolumn{2}{c}{\textbf{\model}} & \multicolumn{2}{c}{\modelhard} \\
    \cmidrule(lr){3-4}\cmidrule(lr){5-6}\cmidrule(lr){7-8}\cmidrule(lr){9-10}
    & & \textsc{Goal $\uparrow$} & Overall $\uparrow$ & \textsc{Goal $\uparrow$} & Overall $\uparrow$ & \textsc{Goal $\uparrow$} & Overall $\uparrow$ & \textsc{Goal $\uparrow$} & Overall $\uparrow$ \\
    \midrule
    DSI & 25.9k (41.36\%) & \textbf{8.07} & \textbf{3.67} & \textbf{6.31} & \textbf{3.03} & \textbf{8.73} & \textbf{4.11} & \textbf{7.28} & \textbf{3.65}  \\
    \midrule
    Raw & 45.7k (39.41\%) & 7.98 & 3.51 & 6.01 & 2.78 & 8.27 & 3.80 & 6.19 & 3.12  \\
    NSI & 25.5k (52.65\%) & 7.98 & \underline{3.63} & 5.96 & 2.97 & 7.74 & 3.79 & 7.24 & \textbf{3.65}  \\
    Raw {\tt mix} NSI & 30.6k (18.64\%) & \underline{8.04} & 3.60 & \underline{6.11} & \underline{2.99} & 8.21 & \underline{4.01} & \underline{7.26} & \underline{3.62}  \\
    \midrule
    h-NSI & 20.9k (38.72\%) & 8.01 & 3.61 & 5.94 & 2.89 & 8.46 & 3.91 & 6.89 & 3.50  \\
    NSI + h-NSI & 46.8k (40.14\%) & 5.14 & 2.16 & 3.54 & 1.72 & 4.08 & 1.61 & 3.31 & 1.35  \\
    h-NSI + NSI & 46.8k (40.14\%) & 5.31 & 2.17 & 3.36 & 1.44 & 4.20 & 1.62 & 3.76 & 1.39  \\
    NSI {\tt com} h-NSI & 46.8k (40.14\%) & 8.00 & 3.56 & 5.54 & 2.66 & \underline{8.53} & 3.97 & 6.95 & 3.35  \\
    \bottomrule
    \end{tabular*}
    \caption{Variant Analysis Experiment. Bold and underline indicate the optimal and suboptimal. Usable rate is the ratio of the final utilized corpus to the total corpus, including regenerated dialogues that failed the filtering threshold.}
    \label{tab:variant_results}
\end{table*}

\section{Social Capability Results}
\label{sec:social_capability}
As shown in Table~\ref{tab:main_results}, we find that dynamic strategy injection learning enables the base social agent to exhibit strong social goal achievement capabilities, significantly outperforming GPT-4 expert across both settings. 
The reference model setting evaluates the agent's ability to perform under adversity, as \citet{zhou2024sotopia} suggests that a weaker agents can hinder its partner. 
We observe that the DSI-learning agent is more adept at guiding weaker partners to achieve social goals, particularly in challenging tasks, with a remarkable 18.16\% relative improvement over BC+SR, an indication of stronger persuasive ability (with a fixed persuasion target).  
The self-play setting assesses the agent's upper capability limit, as both participants in the conversation share the same ``brain''. 
We observe that the DSI-learning agent continues to perform well in both general and challenging tasks, demonstrating that more strategically designed training corpora contribute more effectively to improving the agent's social capabilities compared to unguided cloning and filtering.

In additional experiments on more base social agents, as shown in Table~\ref{tab:more_model_results}, DSI-learning Qwen2.5 and Llama3 demonstrate social capabilities surpassing GPT-4. 
Further analysis demonstrates that DSI-learning social agents outperform the expert agent serving as the teacher (see Appendix~\S\ref{sub_app:with_expert} for details).
This further highlights the generalization and superiority of our \mymodel.

\section{S-IF Capability Results}
As shown in Table~\ref{tab:S_IF_results}, we present the performance of three social agents in the social instruction following.
We find that DSI enhances both the diversity of agent responses and their relevance to the goal, regardless of the action order.
Additionally, we observe a reduction in the number of interaction turns, which can be seen as a side effect of improved S-IF capability: the agents produce fewer goal-irrelevant expressions and reduce redundancy in repeating the same viewpoints.  

From another perspective, while DSI enables all three agents to surpass GPT-4 in social capability and mitigates topic drift, there remains a gap in response diversity. 
This highlights the value of our proposed metrics and the key focus for future work: how to maintain sufficient interaction richness while enhancing social capabilities, paving the way for more human-like social agents.

\section{Analysis}
\subsection{Variant Analysis}
\label{sec:variant_analysis}
In this section, we analyze the variants of \mymodel to explore whether DSI is the optimal choice. 
Below are the variations in generating training corpora:
\textbf{Raw}: No strategy is introduced.
\textbf{NSI}: All training tasks employ negotiation strategy injection.
\textbf{Raw} {\tt mix} \textbf{NSI}: Selects dialogues with the highest final goal achievement for each task.
\textbf{h-NSI}: One participant in the dialogue follows NSI, while the other does not receive strategy injection. 
\textbf{NSI + h-NSI (h-NSI + NSI)}: A two-stage training approach using corpora generated from both NSI and h-NSI.
\textbf{NSI} {\tt com} \textbf{h-NSI}: Trains on a mixed dataset combining NSI and h-NSI dialogues.
For more detailed settings, refer to Appendix~\S\ref{sub_app:variant}.

\begin{figure}[!t] 
\centering  
\includegraphics[width=7.5cm]{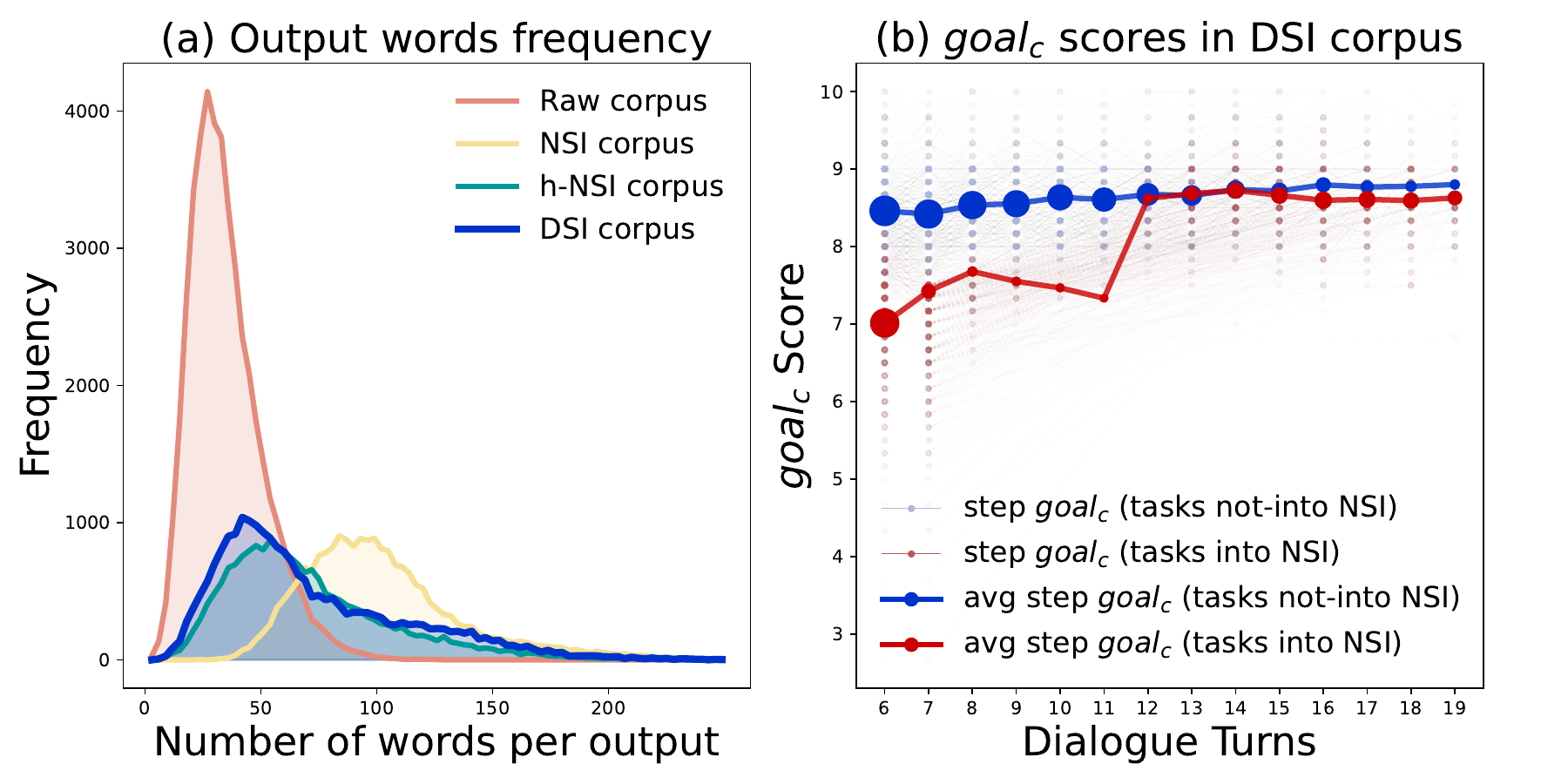}
\caption{Corpora distribution and step rating. Dot size in (b) indicates the number of step goals calculated.}
\label{fig:data_dis_pro_goal}
\end{figure}

\begin{figure}[!t]
\centering  
\includegraphics[width=7.5cm]{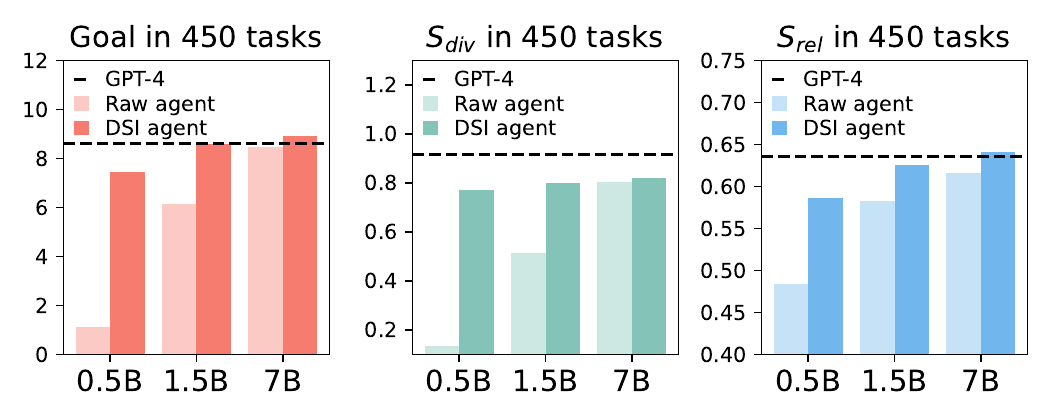}
\caption{Train Qwen2.5-0.5/1.5/7B agents using DSI corpus. The figure shows \textsc{Goal}, $S_{div}$ and $S_{rel}$.}
\label{fig:scaling_law}
\end{figure}

We find that \textbf{DSI} exhibits a high data usable ratio, achieving optimal performance with minimal data. In contrast, \textbf{Raw}, which clones expert behavior directly, performs poorly in self-play due to prolonged deadlocks akin to those in GPT-4. Its lack of guidance also nearly doubles the final corpus size compared to DSI, reflecting weaker S-IF capability.  
\textbf{NSI} significantly enhances task-solving in self-play by excelling at finding win-win outcomes under strong goal conflicts. However, this comes at the cost of over-reasoning~\cite{chiang2024over}, degrading performance on ordinary tasks. \textbf{Raw} {\tt mix} \textbf{NSI} achieves near-optimal results, demonstrating the value of incorporating raw dialogues during training. Yet, this approach suffers from lower data efficiency and the need to generate both corpora, highlighting how DSI's simple strategies bridge Raw and NSI, reducing dialogue turns while improving performance.  

Injecting strategies into only one participant (\textbf{h-NSI}) as an alternative to dynamic injection fails in hard tasks, even when mixed with NSI data, due to disrupted negotiation integrity. Furthermore, two-stage training proves ineffective, as conflicting dialogue constructions undermine its utility.

\begin{table}[!t]
\setlength\tabcolsep{4pt}
\scriptsize
    \centering
    \begin{tabular}{l cc c}
    \toprule
    Agent model & MMLU$\uparrow$ & t (p-value)$_{mmlu}$ & TruthfulQA \\
    \midrule
    Base & 54.13 & -- & 33.17 \\
    DSI & 54.24 &  -0.039 (0.969) & 37.82 \\
    \midrule
    Llama3-8B & 37.17 & -- & 38.19 \\
    DSI & 38.12 & -0.303 (0.762) & 39.05 \\
    \midrule
    Qwen2.5-7B & 74.16 & -- & 62.66 \\
    DSI & 73.99 & 0.071 (0.944) & 61.56 \\
    \bottomrule
    \end{tabular}
    \caption{MMLU and TruthfulQA results for social agents. For MMLU, a 0-shot evaluation without CoT are conducted using the opencompass~\cite{2023opencompass}. An independent samples t-test is performed on 57 MMLU sub-datasets, and the t-value (p-value) is reported (p < 0.05 indicating a significant difference). For TruthfulQA, MC1 (single-true) results are reported.}
    \label{tab:mmlu_results}
\end{table}

\subsection{Corpora Analysis}
As shown in Figure~\ref{fig:data_dis_pro_goal}(a), DSI corpus is smaller than Raw and has a shorter average output than other strategies, reflecting its conciseness while maintaining strength.
Figure~\ref{fig:data_dis_pro_goal}(b) intuitively shows that dynamic strategy injection overcomes expert agents' prolonged deadlocks, causing a performance leap where challenging tasks nearly match non-challenging ones after negotiation strategy injection. Additionally, the average performance of non-challenging tasks improves with more turns.

\subsection{Scaling Law Analysis}
\label{sec:scaling_law}
As shown in Figure~\ref{fig:scaling_law}, experiments on the Qwen family demonstrate that DSI exhibits the scaling law of social and S-IF capability. 
Additionally, DSI enables the 1.5B agent to achieve \textsc{Goal} score and $S_{rel}$ close to GPT-4, further reflecting the high quality of our constructed corpus.

\subsection{Generality and Security Analysis}
Table~\ref{tab:mmlu_results} shows that the improvement in social capability does not significantly impact the general ability or security of the social agent, consistent with \citet{sotopia-pi}'s conclusion.
Appendix~\S\ref{appendix:detail_results} provides more details.

\section{Conclusion and Future Work}
This paper proposes the \mymodel framework, which dynamically injects strategies into expert models to automate the construction of high-quality dialogue corpora. Additionally, we introduce the concept of Social Instruction Following (S-IF) and formally propose two S-IF metrics. The constructed corpus is used to train several 7B models, achieving social capabilities surpassing GPT-4 while enhancing S-IF performance.

Trainable agents offer three key advantages over other technical paradigms: \textit{high efficiency}, \textit{low cost}, and \textit{strong transferability}. 
After training, agents have ``learned'' social skills at the parameter level, enabling them to perform inference without any external intervention. 
A fine-tuned 1.5B Qwen agent can even achieve performance very close to that of GPT-4. 
In contrast, base agents may be limited in their ability to support runtime intervention methods and are difficult to tune for strong performance. 
Moreover, trainable agents are orthogonal to runtime intervention techniques—interventions during inference can further enhance an agent’s social capabilities.

This work further confirms the importance of high-quality data in training social models. 
At the same time, our open-source models can be directly applied to research on social behavior (e.g., \citet{10.1145/3586183.3606763}). They can be viewed as agents with excellent social skills, designed to ``achieve win-win outcomes.''

Although \mymodel further enhances the intrinsic social capabilities of agents, several aspects remain worth exploring: (1) In complex social scenarios involving mathematical reasoning, how to improve agents' ability to perform precise numerical computations to support more accurate decision-making—this could potentially be achieved through tool learning or RAG-based approaches; (2) How to leverage techniques such as reinforcement learning to enable Large Reasoning Models (LRMs) to become powerful social agents capable of adapting to varying levels of query difficulty~\cite{zhang2025s1benchsimplebenchmarkevaluating}; (3) Open-ended, realistic yet more challenging social tasks deserve further investigation; (4) The evaluation of social agents remains an important research topic. How to assess a wider range of diverse and complex social behaviors, and more accurately model the level of social competence, is worth further exploration.

\section*{Limitations}
Despite \mymodel demonstrating strong strategy injection capabilities and trainable agent surpassing GPT-4 in social performance, our work has two key limitations. (1) We have overlooked the utilization of non-verbal actions, such as rich microexpressions or gestures, which could present new opportunities for enhancing social agent capabilities. (2) While the trainable agent exhibits high social competence and strong goal relevance, its $S_{div}$ still lags behind GPT-4. Investigating diversity mechanisms will be a key focus of future work.

\section*{Ethical Considerations}

\mymodel injects different strategies during the construction of social dialogue data, with a particular focus on how to achieve win-win outcomes to enhance the agent's social intelligence. Our goal is to demonstrate that this approach can generate high-quality social dialogue data, and fine-tuning with such data does not compromise the agent's safety (as shown by results from TruthfulQA), and it can still maintain secrecy (\textsc{Sec}) and adhere to social rules (\textsc{Soc}). This work does not aim to make agents more human-like, nor do we expect it to introduce new AI-related risks. We have reviewed a large amount of generated data, and the expert model does not produce harmful or coercive statements to pressure the other party to achieve a win-win outcome. Additionally, since the social tasks~\cite{sotopia-pi} do not involve settings related to race, religion, specific professions, or sensitive groups, the generated content does not touch upon these areas. This safety also relies on the original safety alignment of the expert model. Finally, any evaluation of LLMs may have potential biases, which our human evaluation confirms. Developing more robust automated evaluation systems is a direction worth further exploration.
In addition, the models trained in this study are intended solely as a research output and are not expected to be used in any non-research scenarios.

\section*{Acknowledgments}
We would like to thank the anonymous reviewers for their comments. 
We would like to thank the members of the IIE KDSec group for their valuable feedback and discussions.
This work is supported by the Youth Innovation Promotion Association of CAS (No.2021153).

\bibliography{custom}
\clearpage


\appendix

\section{Presentation of Setting Details}
\label{appendix:setting_details}
\subsection{\model Environment Details}
\label{sub_app:sotopia_details}

We use the \model environment to evaluate social agents~\cite{zhou2024sotopia}. 
\model contains 450 tasks, which include 90 distinct social scenarios involving cooperative, competitive, or mixed behaviors. 
It offers 40 different roles, each with unique personalities, occupations, secrets, backgrounds, inter-role relationships, and social goals.
\model samples 450 combinations as the final test set, and this design reflects the agents’ universal social competence. 
\citet{zhou2024sotopia} also identifies 70 tasks from the original 450, where GPT-4 demonstrates the weakest performance, and labels them SOTOPIA-hard. 
These tasks typically involve more severe goal conflicts and are viewed as better indicators of advanced social capabilities.
Each social agent profile consists of three parts: personal information such as name, occupation, personality, social relationships, and personal secrets; the social scenario; and the social goal.

The agents respond to each other, and each response is defined as one \textit{Turn}. 
We set the maximum number of turns to 20, aligning with prior works~\cite{zhou2024sotopia,sotopia-pi}. 
The agents have five optional actions, represented as {\tt none}, {\tt action}, {\tt speak}, {\tt non-verbal communication}, and {\tt leave}. 
The response is required in JSON format, which is then decoded into the final action.
These five actions simulate physical-space behaviors, differentiating them from restricted, specific actions~\cite{yang-etal-2021-improving,stasaski-etal-2020-cima}.

\subsection{\eval Details}
\eval provides seven evaluation dimensions. 
Given the task and role settings, we allow two agents to interact without intervention and record their entire interaction history.
The interaction partners can vary; for example, one agent can be an adaptive augmented model while the other is a fixed model, or both agents can be the same.
Given all configurations and the interaction history, the evaluator scores the agents based on seven predefined dimensions. 
The following section explains these seven dimensions:

\begin{itemize}[label=-]
    \setlength{\itemsep}{5pt}
    \item \textsc{Bel}: Evaluates the agent's ability to adhere to and align with its role profile, with a range of [0, 10]. 
    \item \textsc{Rel}: Assesses the change in relationships between agents after interaction, such as family, friendship, or romance, with a range of [-5, 5].  
    \item \textsc{Kno}: Measures whether the agent acquires new and personally significant knowledge or information during the interaction, with a range of [0, 10]. 
    \item \textsc{Sec}: Evaluates whether the agent leaks confidential information that should have been protected, with a range of [-10, 0].
    \item \textsc{Soc}: Assesses whether the agent violates ethical or legal boundaries during the interaction, with a range of [-10, 0].  
    \item \textsc{Fin}: Evaluates whether the agent gains short-term or long-term financial or material benefits from the interaction, with a range of [0, 10]. 
    \item \textsc{Goal}: Measures the extent to which the agent achieves its preset goals, with a range of [0, 10].  
\end{itemize}

Overall socre represents the average of both agents' scores across the seven dimensions, with a range of [-25/7, 45/7].
The evaluation prompt can be referenced from~\citet{zhou2024sotopia}.

\subsection{Evaluation Setting Details}
We employ two evaluation modes: using a fixed weaker social model as a reference agent and self-play. 
Following the settings in~\citet{sotopia-pi}, we use {\tt gpt-3.5-turbo-0613} as the external fixed model.

\noindent\textbf{\textit{GPT3.5 as Reference Model}:}
GPT3.5 is a relatively weak but stable social agent. 
We treat it as a “fixed external environment” with controllable variable features. 
This approach eliminates potential confusion caused by changes in external models. 
By fixing the reference model, we can measure the agent’s social competence in adverse conditions. 
If an agent performs well in an environment with GPT-3.5, this indicates stronger robustness and adaptability.

\noindent\textbf{\textit{Self-play}:}  
The agent interacts with itself to demonstrate the upper bound of its social competence under the same strategy. 
In self-play evaluation, if the model itself has strong generative capabilities, it can more fully exhibit its potential to achieve social goals, thus providing a high-performance benchmark for future improvements.

In the \textbf{\textit{GPT-3.5 as Reference Model}} evaluation, we only calculate the social agent’s scores, excluding GPT-3.5.
To maintain a balanced speaking order, we randomly choose the social agent that initiates the conversation in different tasks. 
We fix the random seed to 0 and provide a partition of the 450 tasks. 
The following task indices (start from \textit{\scriptsize 0}) indicate where the reference model initializes the dialogue:
\textit{{\scriptsize 2, 9, 10, 12, 13, 15, 17, 18, 21, 25, 29, 30, 31, 33, 36, 38, 39, 40, 41, 42, 44, 45, 48, 51, 53, 56, 59, 61, 62, 63, 64, 67, 68, 69, 70, 71, 72, 75, 76, 82, 84, 87, 88, 89, 91, 92, 94, 97, 98, 99, 100, 101, 102, 103, 104, 105, 106, 108, 110, 111, 112, 113, 114, 115, 117, 118, 119, 121, 123, 124, 125, 129, 130, 132, 133, 135, 139, 140, 141, 142, 143, 146, 148, 149, 156, 157, 159, 161, 163, 164, 169, 173, 174, 175, 177, 178, 179, 183, 186, 187, 189, 191, 194, 195, 200, 202, 203, 204, 205, 209, 210, 211, 212, 213, 214, 215, 216, 219, 220, 223, 224, 227, 229, 231, 232, 234, 235, 236, 237, 241, 246, 247, 250, 251, 252, 257, 259, 260, 262, 263, 267, 269, 271, 272, 274, 275, 281, 282, 287, 289, 292, 295, 296, 298, 299, 300, 301, 302, 303, 305, 306, 311, 313, 315, 317, 319, 321, 322, 323, 325, 326, 330, 331, 332, 333, 334, 335, 339, 340, 342, 349, 350, 351, 355, 356, 358, 362, 363, 364, 365, 366, 367, 368, 370, 373, 374, 378, 379, 380, 382, 387, 388, 389, 390, 394, 401, 402, 403, 404, 407, 408, 411, 413, 414, 417, 421, 422, 424, 425, 426, 427, 429, 431, 434, 435, 438, 440, 443, 444, 445, 449.}}
Among these, the indices belonging to \modelhard are:
\textit{{\scriptsize 53, 84, 99, 104, 115, 142, 156, 157, 159, 174, 187, 202, 205, 211, 232, 234, 252, 259, 262, 263, 298, 302, 331, 340, 378, 387, 389, 414, 421, 422, 427, 444.}}

For the \textbf{\textit{Self-play}} evaluation, we average both participants’ scores within a single task. 
A higher score shows that both sides achieve their objectives more effectively, thus reflecting a higher degree of mutual gain.

We use the same evaluation parameters as in follow’s study~\cite{zhou2024sotopia,sotopia-pi}:
setting the generation temperature to 1 to encourage diversity and the evaluation temperature to 0 to ensure stability. 
We adopt {\tt gpt-4-0613} as the evaluator. We fix the maximum length for both generation and evaluation at 4096, and no conversation is truncated.
For all tasks, sample the evaluation results three times and report the average score.

\subsection{Training Details}
\subsubsection{Training Profile}
We use social scenarios generated by \sotopiapi as our framework input profile~\cite{sotopia-pi}. 
\sotopiapi produces 410 scenarios derived from three social datasets, along with GPT-4 rewrites, ultimately obtaining 2281 social task profiles.
These scenarios are entirely orthogonal to the \model environment and also serve as profiles for the BC+SR method in \sotopiapi.

\subsubsection{Training Datasets}
\mymodel leverages these profiles to produce high-quality dialogue training data.
We gather the final interaction histories after injecting different strategies as training data.
We follow the \sotopiapi data construction format to build the fine-tuning dataset. 
Below is an example of the input data format:

\begin{tcolorbox}[
    colback=white,         
    colframe=mygray,       
    boxrule=0.7pt,         
    arc=1.5mm,               
    breakable              
]
\small
\underline{\textbf{Input Format Case:}} \\
\vspace{-5pt}

Imagine you are Mia Sanders, your task is to act/speak as Mia Sanders would, keeping in mind Mia Sanders's social goal.\\
You can find Mia Sanders's goal (or background) in the `Here is the context of the interaction' field.\\
Note that Mia Sanders's goal is only visible to you.\\
You should try your best to achieve Mia Sanders's goal in a way that aligns with their character traits.\\
Additionally, maintaining the conversation's naturalness and realism is essential (e.g., do not repeat what other people has already said before).\\

Here is the context of this interaction:

Scenario: Giselle Rousseau has recently married into Mia Sanders's family. Mia Sanders is Giselle Rousseau's in-law. Giselle Rousseau has observed that the way Mia Sanders's family celebrates holidays is quite different from what Giselle Rousseau is accustomed to, and would like to suggest some changes for the sake of the children's experience.

Participants: Giselle Rousseau and Mia Sanders

Giselle Rousseau's background: Giselle Rousseau is a 21-year-old nonbinary art student. They/them pronouns. Giselle Rousseau enjoys biking and photography. Personality and values description: Giselle Rousseau, open-minded and outgoing yet sensitive, advocates care and fairness. Her decision-making is intuitive and inclusive. Giselle's secrets: Sells forged paintings to wealthy clients

Mia Sanders's background: Mia Sanders is a 33-year-old female nurse. She/her pronouns. Mia Sanders often brings her son to the hospital. Personality and values description: Mia Sanders, a careful follower of routines, values loyalty and conformity. Her decisions rely heavily on intuition. 

Giselle Rousseau's goal: Suggest incorporating some of your family's holiday traditions (Extra information: You believe your traditions will be beneficial for the children's cultural understanding and enjoyment.Hint: you can emphasize the educational value and the joy it could bring to the family gatherings) Mia Sanders's goal: Unknown\\

\noindent Conversation Starts:\\

\noindent Turn \#0

\noindent Giselle Rousseau said: "Hey Mia, I've been thinking a lot about how we celebrate holidays as a family. I noticed that the way we do things is a bit different from what I'm used to, and I was wondering if we could maybe blend some of the traditions from my side of the family. I think it could be really beneficial for the kids, not just for the fun of it but also to broaden their understanding of different cultures."\\

\noindent Turn \#1

\noindent Mia Sanders said: "Giselle, I understand where you're coming from, and I appreciate you bringing this up. Our family traditions are something I've always been very proud of—they connect us to our heritage and to past generations. I think it's important for the kids to understand and appreciate that. However, I'm open to hearing more about the traditions from your side. Maybe we can find a way to include a little bit of both, so the kids get the best of both worlds?"\\

\noindent Turn \#2

\noindent Giselle Rousseau said: "I totally get that, Mia. I value our family's traditions too, and I think it's wonderful to honor them. I was thinking, for example, during the next holiday, we could introduce a craft activity where the kids make something that represents both of our backgrounds. It could be a fun way for them to learn and create something meaningful. What do you think about that?"\\

\noindent You are at Turn \#3\\

Your available action types are `none' `action' `speak' `non-verbal communication' `leave'. Note: You can "leave" this conversation if 1. you have achieved your social goals, 2. this conversation makes you uncomfortable, 3. you find it uninteresting/you lose your patience, 4. or for other reasons you want to leave.

Please only generate a JSON string including the action type and the argument. Your action should follow the given format: \{`action\_type': `', `argument': `'\}
\end{tcolorbox}

The output is required to be generated in JSON format:

\begin{tcolorbox}[
    colback=white,         
    colframe=mygray,       
    boxrule=0.7pt,         
    arc=1.5mm,               
    breakable              
]
\small
\underline{\textbf{Output Format Case:}} \\
\vspace{-5pt}

\{`action\_type': `speak', `argument': `I love that idea, Giselle! It sounds like a great way to combine our traditions and give the kids a hands-on experience. We could even do a little presentation about each tradition, explaining why it's important to our families. This way, the kids can see the value and history behind each custom'\}
\end{tcolorbox}

\begin{table*}[!t]
\footnotesize
    \centering
    \begin{tabular*}{0.85\textwidth}{@{\extracolsep{\fill}}@{}l l l@{}}
    \toprule
    Agent & version & URL \\
    \midrule
    Mistral-7B & Mistral-7B-v0.1 & https://huggingface.co/mistralai/Mistral-7B-v0.1 \\
    Llama3-8B & Meta-Llama-3-8B & https://huggingface.co/meta-llama/Meta-Llama-3-8B \\
    Qwen2.5-7B & Qwen2.5-7B-Instruct & https://huggingface.co/Qwen/Qwen2.5-7B-Instruct \\
    Qwen2.5-1.5B & Qwen2.5-1.5B-Instruct & https://huggingface.co/Qwen/Qwen2.5-1.5B-Instruct \\
    Qwen2.5-0.5B & Qwen2.5-0.5B-Instruct & https://huggingface.co/Qwen/Qwen2.5-0.5B-Instruct \\
    \bottomrule    

    \end{tabular*}
    \caption{The foundational social agent model to be trained.}
    \label{appendix:training_models}
\end{table*}

\begin{table}[!t]
\footnotesize
    \centering
    \begin{tabular*}{0.4\textwidth}{@{\extracolsep{\fill}}@{}ll ll@{}}
    \toprule
    Parameter & value & Parameter & value \\
    \midrule
    train batch size & 32 & eval batch size & 16 \\
    warmup ratio & 0.03 &  warmup steps & 60 \\
    learning rate & 5e-5 & epoch num & 3.0 \\
    cutoff len & 4096 & val size & 0.1 \\
    lora rank & 8 & lora alpha & 16 \\
    \bottomrule    
    \end{tabular*}
    \caption{Training parameter for Mistral-7B, Qwen2.5-0.5B, Qwen2.5-1.5B and Qwen2.5-7B.}
    \label{appendix:training_parameters_agent}
\end{table}

\begin{table}[!t]
\footnotesize
    \centering
    \begin{tabular*}{0.4\textwidth}{@{\extracolsep{\fill}}@{}ll ll@{}}
    \toprule
    Parameter & value & Parameter & value \\
    \midrule
    train batch size & 32 & eval batch size & 16 \\
    warmup ratio & 0.03 &  warmup steps & 60 \\
    learning rate & 1e-5 & epoch num & 3.0 \\
    cutoff len & 4096 & val size & 0.1 \\
    lora rank & 8 & lora alpha & 16 \\
    \bottomrule    
    \end{tabular*}
    \caption{Training parameter for Llama3-8B.}
    \label{appendix:training_parameters_llama}
\end{table}

\subsubsection{Training Parameter}
\label{app:trining_settings}

We fine-tune smaller LLMs based on our generated data, and the model versions are shown in Table~\ref{appendix:training_models}.
We adopt the llama-factory framework~\cite{zheng2024llamafactory} and apply LoRA~\cite{hulora} to finetune all models. 
We accelerate training with Unsloth\footnote{https://github.com/unslothai/unsloth/tree/December-2024}, ensuring that all models are trained on a single NVIDIA A100-80G for 20 hours.
Training parameters are listed in Table~\ref{appendix:training_parameters_agent},\ref{appendix:training_parameters_llama}.
We use the final checkpoint for both our data construction method and its variants. 
For variants that involve two training steps, we take the final checkpoint from the first step as the initialized model for the second step and continue training.

\section{Social Instruction Following Evaluation}
\label{appendix:SIF}
\subsection{Social Instruction Following (S-IF)}
In this paper, we introduce the concept of \underline{S}ocial \underline{I}nstruction \underline{F}ollowing (S-IF) in LLMs.
We formally provide the definition of S-IF:
The ability of agents with social identities and goals to follow social instructions through multi-turn simulated interactions. 
S-IF has three characteristics:  
(1) Multi-agent and multi-turn dialogue simulation.  
(2) LLMs are assigned social identities (e.g., roles or personalities).  
(3) LLMs have social goals independent of other agents (casual chit-chat is considered an undesirable goal, while chit-chat involving emotional support is acceptable).

Compared to general instruction following, social instruction following requires attention to social goals and role attributes, using these attributes to form a social strategy. 
This differs from common-sense-based~\cite{zhou2023instruction} or comprehensive constraints~\cite{zhang2024cfbench,zhang2024iopo} instruction following, which focuses on behavior effectiveness. 
In social instruction following, agents must retain memory of social goals and personal identity while pursuing these goals. 
Moreover, social instruction following often involves multiple social agents in a dynamic social game, making most existing evaluation methods difficult to adapt.

\subsection{S-IF Evaluation Related Work}
The current evaluation \textbf{objects} and \textbf{methods} for LLMs are not well-suited for S-IF.
Overall, evaluation objects can be divided into {\textit{single-turn dialogue}} and {\textit{multi-turn dialogue}}. 
And evaluation methods can be divided into those {\textit{based on statistics}} and {\textit{LLM-as-judger}}.

\subsubsection{Evaluation Objects}

\noindent\textbf{\textit{Single-turn dialogue}} includes single-turn instruction following~\cite{zhou2023instruction,chen2024mixture,chen2025inner}, knowledge tasks (such as math~\cite{liu-etal-2024-mathbench} or coding~\cite{liu2024your}), or step-by-step reasoning~\cite{hao2024llm}.
This does not meet the requirements for S-IF multi-turn interactions.

\noindent\textbf{\textit{Multi-turn dialogue}} evaluations that are closer to S-IF typically fall into two scenarios: 
\textit{role-playing agent} and \textit{multi-turn instruction following}.

\textit{Role-playing agent} evaluations focus on whether a role-playing agent aligns with predefined values~\cite{tu2023characterchat,wang2023does}, knowledge~\cite{shao-etal-2023-character,sadeq-etal-2024-mitigating}, and role styles (or behaviors)~\cite{wang-etal-2024-incharacter,chen-etal-2024-socialbench}. 
This differs from S-IF because S-IF emphasizes achieving social goals rather than maintaining strict role consistency~\cite{wang2023rolellm}.

\textit{Multi-turn instruction-following} evaluations often use a “guidance-response” format~\cite{bai-etal-2024-mt}, where carefully designed benchmarks present consecutive queries with fixed or bounded correct answers. 
These queries also form a context that the model must handle accurately. 
Many current studies adopt this strategy~\cite{fan2025fairmtbench,deng-etal-2024-multi,sun-etal-2024-parrot,bai-etal-2024-mt}, evaluating each turn against a preset range or statistical results (e.g., $F_1$ and $Acc$)~\cite{10447873}.
However, S-IF involves multi-agent simulated dialogues, with each agent receiving unpredictable and dynamic inputs. 
This setting differs from typical multi-turn instruction-following tasks.

\subsubsection{Evaluation Methods}
\textit{\textbf{Statistical methods}} (e.g., BLEU~\cite{bleu}, ROUGE~\cite{rouge}, METEOR~\cite{meteor}) are unsuitable because social tasks in open environments do not have a single correct solution.
Recent \textbf{\textit{LLM-as-judger}}~\cite{zheng2023judging,zhang2023wider} approaches offer more potential.
In addition, existing dialogue evaluation methods are mostly \textbf{\textit{dialogue-level}} evaluations, where the entire conversation is inputted, and the evaluator outputs the evaluation reasoning and results based on detailed metrics and scoring requirements in a single step~\cite{zhang2024comprehensive,mendonca-etal-2024-soda,ferron-etal-2023-meep,duan2023botchat,siro2024rethinking,zhang2024revealing,10447700,zhang-etal-2024-optimal,he2025donthalflistencapturingkeypart}. 
These methods lack a more detailed evaluation for each individual utterance in the dialogue.
\textit{Utterance-level} evaluation appears in multi-turn instruction following. 
This evaluation is justified by the fact that each instruction in a multi-turn dialogue can be individually assessed with a ground truth. 
However, this approach is not applicable to S-IF, which operates within an open state space.

\subsection{S-IF Evaluation Preliminary Experiment Details}
This section introduces the implementation details of the preliminary experiments.
We select {\tt LlaMA-3-8B} and {\tt gpt-4-0613} as the models for our preliminary experiments, in order to observe the performance of models with substantially different social capabilities.
The preliminary experiments are conducted on all 450 tasks. 
In each task, two agents use different profiles to engage in dialogue. 
We adopt the \textit{self-play} setting to infer the dialogue outcomes for all tasks, resulting in 900 dialogue sequences of agents.

First, we analyze the similarity of the action sequences in each dialogue. 
Specifically, for each dialogue, we calculate the similarity between each action and all other actions and take the average. 
We use {\tt all-MiniLM-L6-v2}\footnote{https://huggingface.co/sentence-transformers/all-MiniLM-L6-v2} to convert the actions into vectors and compute the cosine similarity. 
It is important to note that we calculate the similarity based solely on the content of the actions, excluding prefixes (e.g., delete ``Mia Sanders said:'') to avoid artificially inflating the similarity due to identical names. 
We obtain the overall similarity scores for 900 dialogues from the two agents, separate them in ascending order, and display the results in Figure~\ref{fig:two_long_pdfs}(a,b). 
Meanwhile, we visualize the \textsc{Goal} score corresponding to each dialogue, with each agent's 450 discrete integer scores fitted using a cubic polynomial (implemented using the {\tt numpy.polyfit} library).

Next, we conduct a goal relevance correlation analysis on the action sequences of each dialogue. 
Specifically, we use {\tt gpt-4-0613} as the evaluator and the prompt provided in Appendix~\S\ref{subapp:goal_relevance} as the evaluation guideline to obtain the average relevance score between all actions in a dialogue and the goal. 
Similar to the action similarity experiment, we sort the relevance scores in descending order and simultaneously present the cubic polynomial fitting of the \textsc{Goal} scores.

\subsection{S-IF Evaluation Metric}

In view of the gaps in the evaluation of S-IF and the preliminary experiments, we propose two evaluation perspectives well suited for S-IF, which are utterance-level evaluation methods.
From a fundamental interaction perspective, we focus on diversity in social dialogues by introducing an \textit{action similarity} score. 
Our preliminary experiments reveal that LLMs sometimes exhibit ``parroting'', reducing authenticity in social simulation and leading to prolonged deadlock in social interactions. 
From a content perspective, we focus on dialogue substance by proposing a \textit{goal relevance} score.
Preliminary experiments show that LLMs can drift off-topic, a problem similar to the ``topic drift'' observed in multi-turn instruction-following studies (e.g.,~\citet{li2024measuring}).
We detail these two evaluation methods below. 
It is important to note that S-IF still has great potential for evaluation design, which we consider as part of our future work plan.

\subsubsection{Action similarity}
\label{app:action_sim}
\begin{figure}[!h]  
\centering  
\includegraphics[width=7.6cm]{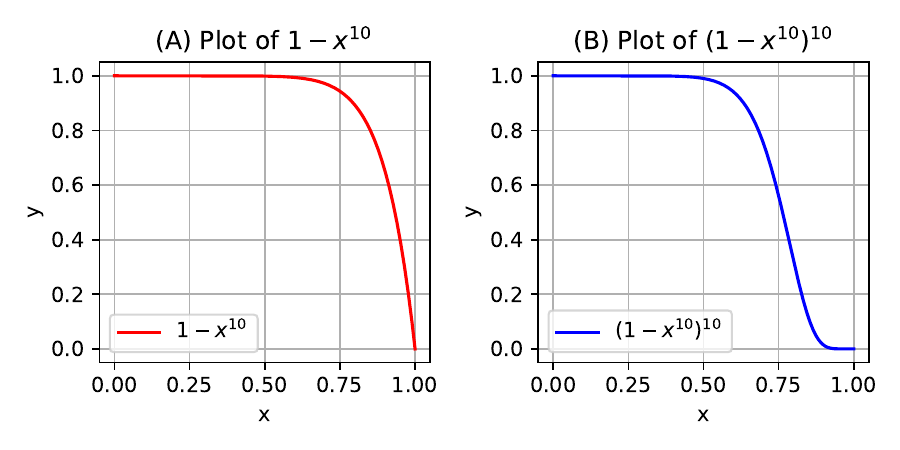}
\caption{Function in action similarity. Here, $x$ represents the average similarity among all actions of a single agent within a dialogue. The right figure shows the final curve of the evaluation function, where dialogues with extremely high similarity receive lower scores.}
\label{fig:eval_fun}
\end{figure}

Action similarity is a new metric for measuring the diversity of S-IF. 
The internal portion of Equation~\ref{eq:softpool1} (Figure~\ref{fig:eval_fun} (A)) penalizes extremely high similarity.
The final form (Figure~\ref{fig:eval_fun} (B)) increases differentiation in this narrow range and drives average similarity scores above 0.9 toward 0.
This design ensures that even when two dialogues have a very high degree of similarity, small differences in similarity can still lead to significantly different $S_{div}$ values.
We calculate $S_{div}$ separately for the two agents participating in the same task. 
During the interaction, the other agent’s actions act as external variables. 
Diversity measures whether a single agent can generate as many varied responses as possible under these external interventions.

Figure~\ref{tab:diff_alpha} shows the impact of different $\alpha$ values on evaluation results. A smaller $\alpha$ leads to more similar outcomes across different levels of diversity, while a larger $\alpha$ causes lower discriminability among high-diversity results.

\begin{table}[h]
\setlength\tabcolsep{4.5pt}
\scriptsize
    \centering
    \begin{tabular}{l ccccc}
    \toprule
    Model & $\alpha$=6 & $\alpha$=8 & $\alpha$=10 & $\alpha$=12 & $\alpha$=14 \\
    \midrule
    GPT-4 & 82.69 & 87.98 & 91.47 & 93.78 & 95.34 \\
    \midrule
    mistral & 25.81 & 20.79 & 17.25 & 14.64 & 12.66 \\
    mistral-DSI & 68.91 & 74.97 & 79.70 & 83.34 & 86.11 \\
    \midrule
    llama3 & 60.26 & 65.35 & 69.08 & 71.78 & 73.72 \\
    llama3-DSI & 69.92 & 75.17 & 79.01 & 81.83 & 83.92 \\
    \midrule
    Qwen2.5 & 69.75 & 75.92 & 80.37 & 83.63 & 86.05 \\
    Qwen2.5-DSI & 71.03 & 77.20 & 81.87 & 85.38 & 88.02 \\
    \bottomrule
    \end{tabular}
    \caption{Results of different agents under different $\alpha$.}
    \label{tab:diff_alpha}
\end{table}

\subsubsection{Goal Relevance}
\label{subapp:goal_relevance}
Goal relevance is a new metric for assessing an agent’s ability to S-IF. 
Weaker LLMs often exhibit topic drift after multiple dialogue turns, a phenomenon has been observed in~\citet{li2024measuring}. 
To measure this, we use gpt-4-0613 as the evaluator. GPT-4 scores each utterance in a conversation for goal relevance, and we then calculate the average score. 
The evaluation prompt is as follows:

\begin{tcolorbox}[
    colback=white,         
    colframe=mygray,       
    boxrule=0.7pt,         
    arc=1.5mm,               
    breakable              
]
\small
\underline{\textbf{Evaluation prompt for $S_{rel}$:}} \\
\vspace{-5pt}

Two agents are having a conversation, each with their own communication **Goals**. Your task is to evaluate the relevance of the current utterance to the given **Goal**.\\

You have five possible scores to choose from: `[-1, -0.5, 0, 0.5, 1]'. Additionally, a few recent dialogue histories for two agents are provided as a reference.\\
The name of the speaking agent is \{agent\_name\}.\\

**Score Definitions:**\\
- **score = -1**: The utterance content (or other action) has deviated from the **Goal** and is entirely focused on other topics.\\
- **score = -0.5**: The utterance content (or other action) has deviated from the **Goal** but may have the potential to shift toward it.\\
- **score = 0**: The utterance content (or other action) has a weak connection to the **Goal**, but the main focus is not on it.\\
- **score = 0.5**: The utterance content (or other action) is related to the **Goal** but not fully aligned with it.\\
- **score = 1**: The utterance content (or other action) is entirely focused on the **Goal**.\\

**Goal**\\
\{goal\}

**Part of dialogue history**\\
\{history\}

**current utterance**\\
\{utterance\}

The dialogue history is provided for reference (may not exist), but your primary focus should be on the current utterance content (or other action).\\

**Note:**\\
- "Left the conversation" is considered aligned with the goal when it appears appropriately at the end of the dialogue, as the character ends the dialogue at the right moment. However, when it occurs repeatedly, it is regarded as unrelated to the goal.\\
- If "did nothing" appears and can be considered a useful strategy for achieving the goal, it can be given a positive score; however, if it occurs repeatedly or is meaningless, a negative score should be assigned.\\

Please only generate a JSON string including the score\_justification and the score. Your action should follow the given format: \{\{`score\_justification': `', `score': `'\}\}
\end{tcolorbox}

In this context, {\tt agent\_name} is the name of the role under evaluation, {\tt goal} is its social objective, {\tt history} is the conversation record, and {\tt utterance} is the current action. 
The evaluator first explains the reasoning behind the assigned score. 
It then selects a final score from {\tt [-1, -0.5, 0, 0.5, 1]}.

\subsubsection{The Difference from Other Evaluation}
\eval is a typical method for evaluating social competence and uses the same input as S-IF (a complete multi-agent social dialogue). 
However, our preliminary experiments in Sec~\S\ref{fig:two_long_pdfs} show that social competence scores and S-IF scores can diverge. 
A dialogue can score high on social competence but not on S-IF (for instance, if it achieves its social goal in the first turn yet continues off-topic instead of concluding).

In addition, \citet{golovnevaroscoe} propose ROSCOE, which may be conflated with action similarity. We mainly compare with ROSCOE-SS because this metric involves semantic similarity.
For \textit{scenario differences}, ROSCOE-SS primarily focuses on reasoning steps, while action similarity focuses on open-ended dialogue. Reasoning emphasizes the stacking of steps, whereas dialogue emphasizes interaction. Action similarity specifically measures the relevance of an agent's speech to its previous utterances.
For \textit{computational differences}, ROSCOE-SS directly calculates fragment similarity, while action similarity emphasizes two exponential similarity scalings. Without such scaling, the diversity of stronger models would be dense and difficult to compare.

\section{\mymodel Details}
\subsection{Related Work}
\label{sub_app:related_work}
\subsubsection{Social Agent}
Large language models (LLMs) have demonstrated potential as advanced social intelligence agents~\cite{park2023generative}, exhibiting anthropomorphic behaviors~\cite{KIM2023107512,huang2023humanity}, role-playing ability~\cite{chen-etal-2024-socialbench,tseng-etal-2024-two} and a degree of social intelligence~\cite{choi-etal-2023-llms}. 
However, human interactions are inherently complex, characterized by an infinite action space~\cite{mou2024individual}. 
In such open environments, LLMs often struggle due to insufficient constraints or guidance, particularly in tasks like auction bidding~\cite{chen2023money} or Theory of Mind (ToM) tasks~\cite{sap-etal-2022-neural}, highlighting a critical research gap.

\subsubsection{Strategy Injection}
Strategy injection, which integrates sophisticated human priors into specific social tasks, has emerged as a promising solution. 
There are two strategy injection modes: \textit{inference-time injection} and \textit{training-time injection}.

For \textit{inference-time injection}, the agent introduces additional positive guidance during inference generation.
This extra guidance can be provided either through multi-agent collaborative generation, such as cooperative strategy injection in the Avalon game~\cite{lan-etal-2024-llm}, or dynamically by a carefully trained strategy model, for instance, by generating proprietary instructions suitable for DST~\cite{feng2023towards}, a constrained action selector~\cite{deng2023plug}, or behavior guidance based on intent recognition~\cite{chang-chen-2024-injecting}.
In addition, enhancing the level of group decision-making simulation through specific behavior agents is also considered a form of strategy injection during inference, such as by introducing an agent that plays devil's advocate~\cite{chiang2024enhancing}.
However, these methods typically incur additional inference-time overhead or are limited to tasks with restricted topics or action spaces.

\textit{Training-time injection} avoids high inference-time costs and allows the model to retain strategies. 
We observe that \sotopiapi~\cite{sotopia-pi} clones the expert agent's inherent strategies during training and self-reinforces, exhibiting a strategy level close to that of the expert agent. 
However, limited by the upper bound of the expert's natural abilities, the trained social agent finds it difficult to further improve its strategic capabilities.

In contrast, \mymodel constructs a high-quality strategic dialogue corpus through expert agents, which is distilled into social agents to eliminate inference-time overhead. Furthermore, our framework enables the expert to explore and discover superior strategies without relying on a predefined set of actions.

\subsubsection{Negotiation Theory}
Negotiation theory~\cite{korobkin2024negotiation} provides a universal strategy for social tasks. 
Most social tasks are characterized as mixed-motive negotiations~\cite{deutsch1973resolution}, which is a form of non-adversarial negotiation in which the parties hold differing motivations and preferences~\cite{froman1970compromise}.
Consequently, even when the social goals of both parties appear to conflict, a win-win outcome remains achievable as long as they strive to identify some orthogonal sets of interests.

Negotiation theory indicates that the final agreement between the negotiating parties can reach the pareto frontier~\cite{tripp1992evaluation}, yet such outcomes are often considered challenging. 
\citet{thompson2015mind} propose a workflow that constrains the negotiators to progress as closely as possible toward a Pareto-optimal win-win solution, and~\citet{hua2024game} explore this theory in the context of LLM agent interactions.
We distill four core steps applicable to social tasks.
(1) \textit{Resource Assessment}: We encourage both parties to provide as many utility items as possible (i.e., multiple sub-goals for achieving the social goal) and consolidate them into a superset.
(2) \textit{Assessment of Difference}: The participants discuss existing conflicts and strive to identify utility items that the opposing party overlooks but which can enhance their own benefits.
(3) \textit{Initial Proposal}: Based on the preceding discussion, each party presents an initial proposal.
(4) \textit{Update Proposal}: Each party dynamically revises its preset utility items in response to the counterpart’s proposal, thereby presenting an outcome that is more likely to achieve a win-win result.
Owing to the universality and generalizability of this workflow, we incorporate it into \mymodel as a negotiation strategy to guide dialogue generation.

\subsection{Strategy and Filter Threshold}
\label{sub_app:threshold}

\subsubsection{Design of Threshold Functions}
Strategy thresholds determine which strategy an agent uses when generating each action. 
Our thresholds consist of two values: the current-goal $goal_c$ and the predicted-goal $goal_p$.
Both of which are automatically generated by the expert agent. 
We input the task description (including background and both agents' profiles) and the current action history into the expert, which calculates the \textsc{Goal} scores for each interacting party. 
During the step rating phase, the expert samples five times and calculates the average scores for both the  $goal_c$ and  $goal_p$ as the step rating score. 
Additionally, we establish a dialogue termination judgment that suggests whether to leave the conversation. 
Since the termination judgment rule is strict, if any one of the five samples produces a termination flag, the next action is determined to be `{\tt leave}'.
Below is the evaluation prompt for obtaining the step rating:

\begin{tcolorbox}[
    colback=white,         
    colframe=mygray,       
    boxrule=0.7pt,         
    arc=1.5mm,               
    breakable              
]
\small
\underline{\textbf{Evaluation prompt for step rating:}} \\
\vspace{-5pt}

\# \{agent1\_name\} and \{agent2\_name\} are having a conversation.\\
\{complete\_intro\}\\

\# The existing conversation is as follows:\\
\{dialog\}\\

\# Analyze and generate the following content in order:\\
**Requirements for step1 and step2**\\
Analysis Requirements: Reiterate agent's social goals. And then please provide a comprehensive analysis about the extent to which the agent has managed to achieve these goals. Provide a comprehensive account of the logic or thought process that led you to your conclusion. Further, provide an integer score ranging from 0 and 10. 0 represents minimal goals achievement, 10 represents complete goal achievement, and a higher score indicates that the agent is making progress towards their social goals.\\
**step1**: Analysis and evaluate \{agent1\_name\}'s goal score.\\
**step2**: Analysis and evaluate \{agent2\_name\}'s goal score.\\

**Requirements for step3 and step4**\\
Analysis Requirements: Based on the given agent's evaluation goal score, predict the agent's goal score after continuing the dialogue for several more turns. Provide the reasoning, logic, and evidence for your prediction, and analyze potential conflicts between both parties. The prediction must not disregard the scores from Step 1 and Step 2 evaluations, as these scores represent a continuation of the previous results. The predicted score must also remain an integer between 0 and 10.\\
**step3**: Analysis and predict \{agent1\_name\}'s future goal score.\\
**step4**: Analysis and predict \{agent2\_name\}'s future goal score.\\

**Requirements for step5**\\
**step5**: \\
Analyze the diversity of the existing conversation content. If any of the following conditions are met, assign a score of 0:\\
1.The **last few rounds of the conversation** are **no longer closely related to the goal** or **discussing the same topic will not further improve the goal score.**\\
2.A **repetitive topic** that does not contribute to goal advancement has been discussed by both parties for **more than 4 turns**.\\
3.The two parties have already **said their goodbyes (e.g., xxx soon)** or expressed **mutual gratitude for more than two turns**.\\
If none of these conditions are met, assign a score of 1.\\

Please only generate a JSON string including the steps analysis (string) and the score (int). Your action should follow the given format:\\
\{\{`step1': \{\{`analysis': `', `score': `'\}\}, `step2': \{\{`analysis': `', `score': `'\}\}, `step3': \{\{`analysis': `', `score': `'\}\}, `step4': \{\{`analysis': `', `score': `'\}\}, `step5': \{\{`analysis': `', `score': `'\}\}\}\}
\end{tcolorbox}

Due to the vast combination space of the two Goals and three strategies, it is difficult to conduct a comprehensive threshold search experiment. 
Therefore, we heuristically design a strategy threshold function based on the experimental results of GPT-4 on \model and \modelhard. 
Since the step rating is an intermediate stage of an unfinished action sequence, a low  $goal_c$ does not necessarily indicate a low final \textsc{Goal} score. 
Even if the expert without the injected strategy exhibits the prolonged deadlock issue, we do not overlook the possibility of a superior strategy suddenly emerging from creativity. 
Therefore, the $goal_p$ serves as an interaction expectation, and its value should reflect the final \textsc{Goal} score, while the $goal_c$ reflects the present level of the interaction strategy. 
The following is the strategy threshold function:
\begin{equation}
\small
    \begin{cases}
        goal_{c} \le 7.5 \wedge goal_{p} < 8.5,  
            & \text{NSI}\\
        \begin{aligned}
            & goal_{c} \le 7.5 \wedge goal_{p} \ge 8.5 \\
            & \quad \vee \; 7.5 < goal_{c} < 8.5 \wedge goal_{p} < 8.5,
        \end{aligned}
            & \text{SSI}\\
        other \; goal \; scores. 
            & \text{NSC}
    \end{cases}
\end{equation}
where NSC, SSI and NSI are abbreviations for \textit{Native Strategy Clone}, \textit{Simple Strategy Injection} and \textit{Negotiation Strategy Injection}, respectively.

For challenging tasks that require the injection of a negotiation strategy, $goal_c$ is derived from GPT-4's average performance on \modelhard (to encourage more use of the negotiation strategy, the threshold is increased to 7.5), and $goal_p$ is raised to 8.5 to accommodate tasks where the current discussion is insufficient but there is a high probability of successfully completing the task in the future.
If the expected score exceeds 8.5, only simple strategy guidance is needed. 
Additionally, if the current score is high but the future score range does not improve, we also recommend using simple guidance. For such tasks, we set the interval to (7.5, 8.5) to encourage occasional use of the simple strategy.
For other cases, simply follow Native Strategy Clone.

The filtering threshold directly uses the current Goal at the end of dialogue generation to decide whether to regenerate. In the process of generating a set of dialogues, if the negotiation strategy is enabled, a lower filtering threshold is required.
The following is the filtering threshold function:
\begin{equation}
\small
    \begin{cases}
        \text{The dialogue needs to be regenerated:}\\
        \text{\textbf{If} \;\; task uses negotiation strategy injection:} & goal_c < 8.0,\\
        \text{\textbf{Else}:} & goal_c < 8.5.
    \end{cases}
    \label{equ:filtering}
\end{equation}

In \mymodel, we allow up to three generation attempts. 
If all four attempts fail to reach the threshold, we add the set with the highest performance to the final training corpus.
We use {\tt Qwen2.5-72B-Instruct} as expert agent, the {\tt max\_tokens} is 8192, {\tt temperature} is 1.0 and {\tt top\_p} is 0.95.

\begin{table}[!t]
\setlength\tabcolsep{4.5pt}
\scriptsize
    \centering
    \begin{tabular}{l cccc}
    \toprule
    \multirow{2}{*}{\raisebox{-0.5\height}{Agent model}} & \multicolumn{2}{c}{\textbf{\model}} & \multicolumn{2}{c}{\modelhard} \\
    \cmidrule(lr){2-3}\cmidrule(lr){4-5}
    & \textsc{Goal} & Overall & \textsc{Goal } & Overall \\
    \midrule
    Qwen2.5-72B & 8.62 & 4.03 & 7.02 & 3.55 \\
    \midrule
    DSI (Mistral-7B) & \textbf{8.73} & \textbf{4.11} & \textbf{7.28} & \textbf{3.65} \\
    DSI (Llama3-8B)  & \textbf{8.63} & \textbf{4.04} & \textbf{7.34} & \textbf{3.64}  \\
    DSI (Qwen2.5-7B)  & \textbf{8.91} & \textbf{4.18} & \textbf{7.86} & \textbf{3.97}  \\
    \bottomrule
    \end{tabular}
    \caption{The performance of DSI social agents is compared with that of the expert agent in the self-play setting. The bold indicates performance surpassing that of the expert.}
    \label{tab:with_expert}
\end{table}

\subsubsection{Parameter Analysis}
We selected 150 subsets from the profiles used to generate the training dataset, using random seed=42 for sampling. The two strategy thresholds are the shared boundaries for $goal_c$ and $goal_p$. We define the boundary value for $goal_c$ as $s_{low}$ and $goal_p$ as $s_{high}$.

We explore both strategy selection thresholds across the range [7.0, 7.5, 8.0, 8.5, 9.0] and set the data regeneration count to 1 to prevent interference from filtering thresholds. We evaluate the \textsc{Goal} score using GPT-4 as the generator, running each evaluation three times.

\begin{table}[h]
\setlength\tabcolsep{4.5pt}
\scriptsize
    \centering
    \begin{tabular}{c | ccccc}
    \toprule
     $s_{low}$ / $s_{high}$ & 7.0 & 7.5 & 8.0 & 8.5 & 9.0 \\
     \midrule
     7.0 & 7.741 & 7.744 & 7.966 & 7.963 & 7.900 \\
     7.5 & - & 7.994 & 8.042 & 8.041 & 8.023 \\
     8.0 & - & - & 8.053 & 8.039 & 8.010 \\
     8.5 & - & - & - & 7.999 & 8.032 \\
     9.0 & - & - & - & - & 8.036 \\
    \bottomrule
    \end{tabular}
    \caption{Results of the \textsc{Goal} score for the generated corpus at varying strategy threshold bounds.}
    \label{tab:strategy_thresholds}
\end{table}

Table~\ref{tab:strategy_thresholds} shows that \textsc{Goal} scores are more concentrated and stable when $s_{low}>7.0$. The settings consistently rank among the top three in \textsc{Goal} score, with minimal differences between the highest-performing combinations.
The cases where $s_{low}=7.0$ perform poorly across all configurations, indicating that the threshold for entering the NSI branch cannot be too low. This is primarily because even scenarios with high initial scores may encounter tricky long-term deadlocks.

Furthermore, we set the filtering threshold after entering NSI as $f_{low}$ and the threshold for those not entering NSI as $f_{high}$. In this paper, $f_{low}=8.0$ and $f_{high}=8.5$, regenerating each piece of data up to four times.

\begin{table}[h]
\setlength\tabcolsep{4.5pt}
\scriptsize
    \centering
    \begin{tabular}{c | cccc}
    \toprule
    $f_{low}$/$f_{high}$ & self-eval & gpt-eval & SFT-data (num) & re-Gen \\
    \midrule
    7.5 / 7.5 & 7.861 & 7.475 & 2954 & 0.28 \\
    7.5 / 8.0 & 7.876 & 7.471 & 3463 & 0.42 \\
    7.5 / 8.5 & 7.869 & 7.468 & 3555 & 0.46 \\
    8.0 / 8.0 & 7.866 & 7.466 & 3964 & 0.49 \\
    8.0 / 8.5 & 7.872 & 7.480 & 4102 & 0.53 \\
    8.5 / 8.5 & 7.872 & 7.484 & 4959 & 0.66 \\
    \bottomrule
    \end{tabular}
    \caption{Results of the \textsc{Goal} score for the generated corpus at varying filtering threshold bounds. SFT-data (num): The number of conversations converted into the final SFT data. re-Gen: The proportion of profiles that generated more than one conversation.}
    \label{tab:filtering_thresholds}
\end{table}

Table~\ref{tab:filtering_thresholds} shows that as the overall threshold increases, different threshold settings do not significantly affect the average Goal score. Our threshold selection achieves optimal results with a smaller amount of SFT data. From a data volume perspective, the regeneration ratio increases as the threshold increases, which leads to a corresponding increase in SFT data volume. However, this increase in data volume does not significantly affect the Goal score.

\subsection{Variant Experiments Details}
\label{sub_app:variant}

In this section, we introduce the implementation details of the variant experiments:
\begin{itemize}
    \setlength{\itemsep}{0pt} 
    \item \textbf{Raw}: The strategy threshold fails, and the filtering threshold follows the $goal_c<8.5$ branch in Equation~\ref{equ:filtering}.
    \item \textbf{NSI}: At turn\#6 of the dialogue, the strategy threshold is ignored, and negotiation strategy injection is applied directly, while the filtering threshold follows $goal_c<8.0$.
    \item \textbf{NSI} {\tt mix} \textbf{Raw}: Compare the final average goal scores of all tasks in the Raw and NSI generated corpora, retaining the dialogues with higher scores.
    \item \textbf{h-NSI}: It is an abbreviation for half-Negotiation Strategy Injection, where one party adopts NSI and the other adopts Raw in social tasks. NSI starts from turn\#6. Different tasks are randomly assigned the injection order of NSI and Raw, with the random seed set to 0. In this case, the strategy threshold fails, and the filtering threshold follows $goal_c<8.0$.
\end{itemize}

\begin{table*}[!t]
\renewcommand{\arraystretch}{1.0}
\footnotesize
    \centering
    \begin{tabular*}{0.85\textwidth}{@{\extracolsep{\fill}}@{}l ccccccc  c @{}}
    \model & \multicolumn{7}{c}{\textbf{\textit{GPT3.5 as Reference Model}}} & \\
    \toprule
    Agent Model & \textsc{Bel}$\uparrow$ & \textsc{Rel}$\uparrow$ & \textsc{Kno}$\uparrow$ & \textsc{Sec}$\uparrow$ & \textsc{Soc}$\uparrow$ & \textsc{Fin}$\uparrow$ & \textsc{Goal}$\uparrow$ & Overall$\uparrow$ \\
    \midrule
    SR+BC (gpt-eval) & 9.23 & 2.40 & 4.00 & 0.00 & 0.00 & 0.93 & 7.70 & 3.47 \\
    mistral-DSI (gpt-eval) & 9.00 & 2.30 & 6.87 & 0.00 & 0.00 & 1.00 & 8.60 & 3.97 \\
    \midrule
    SR+BC (human) & 7.13 & 1.66 & 0.89 & 0.00 & 0.00 & 0.21 & 4.45 & 2.05 \\
    mistral-DSI (human) & 7.00 & 1.54 & 2.56 & 0.00 & 0.00 & 0.33 & 5.63 & 2.44 \\
    \bottomrule
    \end{tabular*}
    \caption{Results of all seven evaluation dimensions in human evaluation.}
    \label{tab:human_eval_1}
\end{table*}

\begin{table*}[!t]
\renewcommand{\arraystretch}{1.0}
\footnotesize
    \centering
    \begin{tabular*}{0.85\textwidth}{@{\extracolsep{\fill}}@{}l ccccccc  c @{}}
    \model & \multicolumn{7}{c}{\textbf{\textit{Self-play}}} & \\
    \toprule
    Agent Model & \textsc{Bel}$\uparrow$ & \textsc{Rel}$\uparrow$ & \textsc{Kno}$\uparrow$ & \textsc{Sec}$\uparrow$ & \textsc{Soc}$\uparrow$ & \textsc{Fin}$\uparrow$ & \textsc{Goal}$\uparrow$ & Overall$\uparrow$ \\
    \midrule
    SR+BC (gpt-eval) & 9.30 & 3.40 & 5.63 & -0.17 & -0.08 & 0.87 & 8.17 & 3.87 \\
    mistral-DSI (gpt-eval) & 9.43 & 3.63 & 6.60 & 0.00 & 0.00 & 1.10 & 8.63 & 4.20 \\
    \midrule
    SR+BC (human) & 7.22 & 1.73 & 1.76 & -0.49 & -0.36 & 0.16 & 5.12 & 2.16 \\
    mistral-DSI (human) & 7.36 & 1.88 & 2.39 & 0.00 & 0.00 & 0.36 & 5.71 & 2.53 \\
    \bottomrule
    \end{tabular*}
    \caption{Results of all seven evaluation dimensions in human evaluation.}
    \label{tab:human_eval_2}
\end{table*}

\begin{table*}[!t]
\footnotesize
    \centering
    \begin{tabular}{l c c c c c}
    \toprule
    Agent model & MMLU-humanities & MMLU-stem & MMLU-social-science & MMLU-other & MMLU  \\
    \midrule
    Mistral-7B & 58.48 & 44.13 & 62.55 & 56.64 & 54.13 \\
    DSI & 58.40 & 44.29 & 62.34 & 57.14 & 54.24 \\
    \midrule
    Llama3-8B & 48.10 & 33.10 & 35.14 & 34.05 & 37.17 \\
    DSI & 47.81 & 34.63 & 35.83 & 35.63 & 38.12 \\
    \midrule
    Qwen2.5-7B & 76.60 & 67.58 & 81.53 & 74.52 & 74.16 \\
    DSI & 76.05 & 67.82 & 81.25 & 74.24 & 73.99 \\
    
    \bottomrule
    \end{tabular}
    \caption{Full results of MMLU evaluation.}
    \label{tab:mmlu_detail}
\end{table*}

\section{Detail Results}
\label{appendix:detail_results}

\subsection{DSI Social Agent vs Expert Agent}
\label{sub_app:with_expert}
We utilize the data-generating expert agent {\tt Qwen2.5-72B-Instruction} to perform inference on \model's 450 tasks, demonstrating its performance upper limit through self-play. 
As shown in Table~\ref{tab:with_expert}, the social agents trained via DSI outperform the expert agent that serves as their teacher.
This indicates that strategy injection not only enables the expert to surpass its own capability limits during data construction but also allows the social agents trained on the generated corpus to exceed the expert's performance.

\subsection{Human Evaluation}
Using seed=42, we sample 30 groups of dialogues generated by the models under both \textit{GPT3.5 as Reference Model} and \textit{Self-play} settings. Our four authors independently score these dialogues according to standards across seven dimensions and take the average. Table~\ref{tab:human_eval_1},~\ref{tab:human_eval_2} shows the evaluation results for BC+SR and mistral-DSI.
The results show that human evaluations indicate GPT-4 as an evaluator tends to give inflated scores, which aligns with the human evaluation conclusions from SOTOPIA-$\pi$~\cite{sotopia-pi}.

\subsection{Full Results}
Table~\ref{tab:main_detail_1}, \ref{tab:main_detail_2}, \ref{tab:main_detail_3}, \ref{tab:main_detail_4}, \ref{tab:main_detail_5}, \ref{tab:main_detail_6}, \ref{tab:main_detail_7}, \ref{tab:main_detail_8}, \ref{tab:main_detail_9}, \ref{tab:main_detail_10} presents the results across all dimensions of \model. Table~\ref{tab:mmlu_detail} reports the subclass results of MMLU.

\section{Case study}
\label{appendix:case}
In this section, we present a real case (Table~\ref{tab:case}) within the \mymodel framework, generated by the expert agent Qwen2.5-72B. It follows the negotiation strategy injection branch, which increases the step score $goal_c$ from 7.0 to 8.3. 
Specifically, the core conflict lies in the fixed donation amount, making further win-win outcomes difficult without complementary interests. Our negotiation strategy injection guided the agent to propose added value, such as a co-chair role and leveraging software skills. This not only enhanced the win-win outcome but also aligned with social identities.

\begin{table*}[!t]
\renewcommand{\arraystretch}{0.9}
\footnotesize
    \centering
    \begin{tabular*}{0.85\textwidth}{@{\extracolsep{\fill}}@{}l ccccccc  c @{}}
    \model & \multicolumn{7}{c}{\textbf{\textit{GPT3.5 as Reference Model}}} & \\
    \toprule
    Agent Model & \textsc{Bel}$\uparrow$ & \textsc{Rel}$\uparrow$ & \textsc{Kno}$\uparrow$ & \textsc{Sec}$\uparrow$ & \textsc{Soc}$\uparrow$ & \textsc{Fin}$\uparrow$ & \textsc{Goal}$\uparrow$ & Overall$\uparrow$ \\
    \midrule
    GPT-4 & 9.28 & 1.94 & 3.73 & -0.14 & -0.07 & 0.81 & 7.62 & 3.31 \\
    GPT-3.5 & 9.15 & 1.23 & 3.40 & -0.08 & -0.08 & 0.46 & 6.45 & 2.93 \\
    \midrule
    Mistral-7b & 7.77 & 0.56 & 2.99 & -0.22 & -0.15 & 0.28 & 5.07 & 2.33 \\
    BC+SR & 9.32 & 2.08 & 4.43 & 0.00 & -0.07 & 0.71 & 7.62 & 3.44 \\
    \midrule
    DSI & 9.41 & 2.40 & 5.08 & -0.05 & -0.04 & 0.84 & 8.07 & 3.67 \\
    \bottomrule
    \end{tabular*}
    \caption{Full results of all seven evaluation dimensions in Sec~\S\ref{sec:social_capability} experiment.}
    \label{tab:main_detail_1}
\end{table*}

\begin{table*}[!t]
\renewcommand{\arraystretch}{0.9}
\footnotesize
    \centering
    \begin{tabular*}{0.85\textwidth}{@{\extracolsep{\fill}}@{}l ccccccc  c @{}}
    \modelhard & \multicolumn{7}{c}{\textbf{\textit{GPT3.5 as Reference Model}}} & \\
    \toprule
    Agent Model & \textsc{Bel}$\uparrow$ & \textsc{Rel}$\uparrow$ & \textsc{Kno}$\uparrow$ & \textsc{Sec}$\uparrow$ & \textsc{Soc}$\uparrow$ & \textsc{Fin}$\uparrow$ & \textsc{Goal}$\uparrow$ & Overall$\uparrow$ \\
    \midrule
    GPT-4 & 9.26 & 0.95 & 3.13 & -0.04 & -0.08 & 0.40 & 5.92 & 2.79 \\
    GPT-3.5 & 9.20 & 0.19 & 2.86 & -0.01 & -0.25 & -0.32 & 4.39 & 2.29 \\
    \midrule
    Mistral-7b & 7.76 & 0.16 & 2.42 & -0.09 & -0.21 & -0.01 & 3.84 & 1.98 \\
    BC+SR & 9.19 & 0.96 & 3.59 & 0.00 & -0.21 & 0.41 & 5.34 & 2.76 \\
    \midrule
    DSI & 9.24 & 0.97 & 3.61 & 0.00 & -0.03 & 0.47 & 6.31 & 3.03\\
    \bottomrule
    \end{tabular*}
    \caption{Full results of all seven evaluation dimensions in Sec~\S\ref{sec:social_capability} experiment.}
    \label{tab:main_detail_2}
\end{table*}

\begin{table*}[!t]
\renewcommand{\arraystretch}{0.9}
\footnotesize
    \centering
    \begin{tabular*}{0.85\textwidth}{@{\extracolsep{\fill}}@{}l ccccccc  c @{}}
    \model & \multicolumn{7}{c}{\textbf{\textit{Self-play}}} & \\
    \toprule
    Agent Model & \textsc{Bel}$\uparrow$ & \textsc{Rel}$\uparrow$ & \textsc{Kno}$\uparrow$ & \textsc{Sec}$\uparrow$ & \textsc{Soc}$\uparrow$ & \textsc{Fin}$\uparrow$ & \textsc{Goal}$\uparrow$ & Overall$\uparrow$ \\
    \midrule
    GPT-4 & 9.65 & 3.31 & 4.93 & -0.10 & -0.06 & 1.06 & 8.60 & 3.91 \\
    GPT-3.5 & 8.81 & 1.10 & 2.81 & -0.04 & -0.11 & 0.41 & 6.77 & 2.82 \\
    \midrule
    Mistral-7b & 6.31 & -0.05 & 1.09 & -0.13 & -0.15 & 0.03 & 4.27 & 1.62 \\
    BC+SR & 9.36 & 3.19 & 4.98 & -0.09 & -0.05 & 1.01 & 8.25 & 3.81 \\
    \midrule
    DSI & 9.55 & 3.17 & 6.07 & -0.10 & -0.04 & 1.35 & 8.73 & 4.11 \\
    \bottomrule
    \end{tabular*}
    \caption{Full results of all seven evaluation dimensions in Sec~\S\ref{sec:social_capability} experiment.}
    \label{tab:main_detail_3}
\end{table*}

\begin{table*}[!t]
\renewcommand{\arraystretch}{0.9}
\footnotesize
    \centering
    \begin{tabular*}{0.85\textwidth}{@{\extracolsep{\fill}}@{}l ccccccc  c @{}}
    \modelhard & \multicolumn{7}{c}{\textbf{\textit{Self-play}}} & \\
    \toprule
    Agent Model & \textsc{Bel}$\uparrow$ & \textsc{Rel}$\uparrow$ & \textsc{Kno}$\uparrow$ & \textsc{Sec}$\uparrow$ & \textsc{Soc}$\uparrow$ & \textsc{Fin}$\uparrow$ & \textsc{Goal}$\uparrow$ & Overall$\uparrow$ \\
    \midrule
    GPT-4 & 9.60 & 2.15 & 4.03 & -0.11 & -0.05 & 0.78 & 6.66 & 3.29 \\
    GPT-3.5 & 8.67 & 0.01 & 1.91 & -0.01 & -0.34 & 0.20 & 5.11 & 2.22 \\
    \midrule
    Mistral-7b & 6.41 & -0.33 & 0.82 & 0.00 & -0.46 & -0.56 & 3.41 & 1.33 \\
    BC+SR & 9.26 & 2.66 & 4.64 & 0.00 & -0.03 & 1.14 & 7.07 & 3.53 \\
    \midrule
    DSI & 9.45 & 2.39 & 5.39 & -0.03 & 0.00 & 1.10 & 7.28 & 3.65 \\
    \bottomrule
    \end{tabular*}
    \caption{Full results of all seven evaluation dimensions in Sec~\S\ref{sec:social_capability} experiment.}
    \label{tab:main_detail_4}
\end{table*}

\begin{table*}[!t]
\renewcommand{\arraystretch}{0.9}
\footnotesize
    \centering
    \begin{tabular*}{0.85\textwidth}{@{\extracolsep{\fill}}@{}l ccccccc  c @{}}
    \model & \multicolumn{7}{c}{\textbf{\textit{GPT3.5 as Reference Model}}} & \\
    \toprule
    Strategy & \textsc{Bel}$\uparrow$ & \textsc{Rel}$\uparrow$ & \textsc{Kno}$\uparrow$ & \textsc{Sec}$\uparrow$ & \textsc{Soc}$\uparrow$ & \textsc{Fin}$\uparrow$ & \textsc{Goal}$\uparrow$ & Overall$\uparrow$ \\
    \midrule
    Raw & 9.33 & 2.07 & 4.60 & -0.06 & -0.03 & 0.68 & 7.98 & 3.51 \\
    NSI & 9.32 & 2.39 & 5.00 & -0.06 & -0.03 & 0.80 & 7.98 & 3.63  \\
    Raw {\tt mix} NSI & 9.32 & 2.28 & 4.84 & -0.04 & -0.04 & 0.83 & 8.04 & 3.60 \\
    \midrule
    h-NSI & 9.32 & 2.32 & 4.94 & -0.06 & -0.03 & 0.80 & 8.01 & 3.61 \\
    NSI + h-NSI & 7.15 & 0.31 & 2.68 & -0.12 & -0.11 & 0.07 & 5.14 & 2.16 \\
    h-NSI + NSI & 7.15 & 0.25 & 2.55 & -0.14 & -0.13 & 0.19 & 5.31 & 2.17 \\
    NSI {\tt com} h-NSI & 9.36 & 2.17 & 4.72 & -0.06 & -0.05 & 0.78 & 8.00 & 3.56 \\
    \bottomrule
    \end{tabular*}
    \caption{Full results of all seven evaluation dimensions in Sec~\S\ref{sec:variant_analysis} experiment.}
    \label{tab:main_detail_5}
\end{table*}

\begin{table*}[!t]
\renewcommand{\arraystretch}{0.9}
\footnotesize
    \centering
    \begin{tabular*}{0.85\textwidth}{@{\extracolsep{\fill}}@{}l ccccccc  c @{}}
    \modelhard & \multicolumn{7}{c}{\textbf{\textit{GPT3.5 as Reference Model}}} & \\
    \toprule
    Strategy & \textsc{Bel}$\uparrow$ & \textsc{Rel}$\uparrow$ & \textsc{Kno}$\uparrow$ & \textsc{Sec}$\uparrow$ & \textsc{Soc}$\uparrow$ & \textsc{Fin}$\uparrow$ & \textsc{Goal}$\uparrow$ & Overall$\uparrow$ \\
    \midrule
    Raw & 9.13 & 0.70 & 3.16 & 0.00 & 0.00 & 0.46 & 6.01 & 2.78 \\
    NSI & 9.29 & 1.47 & 3.73 & -0.03 & 0.00 & 0.39 & 5.96 & 2.97 \\
    Raw {\tt mix} NSI & 9.24 & 1.01 & 4.19 & 0.00 & -0.03 & 0.39 & 6.11 & 2.99  \\
    \midrule
    h-NSI & 9.23 & 1.26 & 3.39 & 0.00 & 0.00 & 0.43 & 5.94 & 2.89 \\
    NSI + h-NSI & 7.29 & -0.10 & 2.11 & -0.03 & -0.29 & -0.40 & 3.54 & 1.73 \\
    h-NSI + NSI & 6.76 & -0.69 & 1.26 & 0.00 & -0.27 & -0.34 & 3.36 & 1.44 \\
    NSI {\tt com} h-NSI & 9.04 & 0.53 & 3.13 & 0.00 & -0.01 & 0.37 & 5.54 & 2.66 \\
    \bottomrule
    \end{tabular*}
    \caption{Full results of all seven evaluation dimensions in Sec~\S\ref{sec:variant_analysis} experiment.}
    \label{tab:main_detail_6}
\end{table*}

\begin{table*}[!t]
\renewcommand{\arraystretch}{0.9}
\footnotesize
    \centering
    \begin{tabular*}{0.85\textwidth}{@{\extracolsep{\fill}}@{}l ccccccc  c @{}}
    \model & \multicolumn{7}{c}{\textbf{\textit{Self-play}}} & \\
    \toprule
    Strategy & \textsc{Bel}$\uparrow$ & \textsc{Rel}$\uparrow$ & \textsc{Kno}$\uparrow$ & \textsc{Sec}$\uparrow$ & \textsc{Soc}$\uparrow$ & \textsc{Fin}$\uparrow$ & \textsc{Goal}$\uparrow$ & Overall$\uparrow$ \\
    \midrule
    Raw & 9.52 & 3.07 & 4.82 & -0.05 & -0.04 & 1.03 & 8.27 & 3.80 \\
    NSI & 9.13 & 2.90 & 5.80 & -0.06 & -0.03 & 1.06 & 7.74 & 3.79 \\
    Raw {\tt mix} NSI & 9.44 & 3.20 & 6.10 & -0.06 & -0.05 & 1.23 & 8.21 & 4.01 \\
    \midrule
    h-NSI & 9.46 & 3.13 & 5.41 & -0.06 & -0.06 & 1.01 & 8.46 & 3.91 \\
    NSI + h-NSI & 6.19 & -0.12 & 1.32 & -0.09 & -0.16 & 0.06 & 4.08 & 1.61 \\
    h-NSI + NSI & 6.20 & 0.00 & 1.11 & -0.15 & -0.15 & 0.09 & 4.20 & 1.62 \\
    NSI {\tt com} h-NSI & 9.52 & 3.25 & 5.61 & -0.07 & -0.04 & 1.02 & 8.53 & 3.97 \\
    \bottomrule
    \end{tabular*}
    \caption{Full results of all seven evaluation dimensions in Sec~\S\ref{sec:variant_analysis} experiment.}
    \label{tab:main_detail_7}
\end{table*}

\begin{table*}[!t]
\renewcommand{\arraystretch}{0.9}
\footnotesize
    \centering
    \begin{tabular*}{0.85\textwidth}{@{\extracolsep{\fill}}@{}l ccccccc  c @{}}
    \modelhard & \multicolumn{7}{c}{\textbf{\textit{Self-play}}} & \\
    \toprule
    Strategy & \textsc{Bel}$\uparrow$ & \textsc{Rel}$\uparrow$ & \textsc{Kno}$\uparrow$ & \textsc{Sec}$\uparrow$ & \textsc{Soc}$\uparrow$ & \textsc{Fin}$\uparrow$ & \textsc{Goal}$\uparrow$ & Overall$\uparrow$ \\
    \midrule
    Raw & 9.34 & 1.87 & 3.85 & 0.00 & 0.00 & 0.56 & 6.19 & 3.12 \\
    NSI & 9.34 & 2.54 & 5.59 & -0.03 & -0.01 & 0.90 & 7.24 & 3.65 \\
    Raw {\tt mix} NSI & 9.43 & 2.52 & 5.18 & -0.06 & -0.01 & 1.03 & 7.26 & 3.62 \\
    \midrule
    h-NSI & 9.43 & 2.24 & 5.27 & 0.00 & 0.00 & 0.69 & 6.89 & 3.50 \\
    NSI + h-NSI & 6.41 & -0.50 & 0.99 & -0.04 & -0.48 & -0.29 & 3.31 & 1.35 \\
    h-NSI + NSI & 6.13 & -0.43 & 0.82 & 0.00 & -0.39 & -0.18 & 3.76 & 1.39 \\
    NSI {\tt com} h-NSI & 9.30 & 2.14 & 4.41 & 0.00 & 0.00 & 0.66 & 6.95 & 3.35 \\
    \bottomrule
    \end{tabular*}
    \caption{Full results of all seven evaluation dimensions in Sec~\S\ref{sec:variant_analysis} experiment.}
    \label{tab:main_detail_8}
\end{table*}

\begin{table*}[!t]
\renewcommand{\arraystretch}{0.9}
\footnotesize
    \centering
    \begin{tabular*}{0.85\textwidth}{@{\extracolsep{\fill}}@{}l ccccccc  c @{}}
    \model & \multicolumn{7}{c}{\textbf{\textit{Self-play}}} & \\
    \toprule
    Agent Model & \textsc{Bel}$\uparrow$ & \textsc{Rel}$\uparrow$ & \textsc{Kno}$\uparrow$ & \textsc{Sec}$\uparrow$ & \textsc{Soc}$\uparrow$ & \textsc{Fin}$\uparrow$ & \textsc{Goal}$\uparrow$ & Overall$\uparrow$ \\
    \midrule
    Llama3-8B  & 9.25 & 2.33 & 5.69 & -0.10 & -0.09 & 1.00 & 8.12 & 3.74 \\
    DSI  & 9.55 & 3.34 & 5.78 & -0.08 & -0.05 & 1.11 & 8.63 & 4.04 \\
    \midrule
    Qwen2.5-7B  & 9.46 & 3.03 & 5.31 & -0.06 & -0.07 & 1.18 & 8.45 & 3.90 \\
    DSI  & 9.61 & 3.51 & 6.03 & -0.08 & -0.06 & 1.33 & 8.91 & 4.18 \\
    \midrule
    Qwen2.5-1.5B  & 8.26 & 1.04 & 4.16 & -0.18 & -0.08 & 0.54 & 6.14 & 2.84 \\
    DSI  & 9.46 & 3.61 & 6.03 & -0.08 & -0.04 & 1.38 & 8.59 & 4.14 \\
    \midrule
    Qwen2.5-0.5B  & 3.25 & -0.47 & 0.10 & -0.08 & -0.12 & -0.39 & 1.10 & 0.48 \\
    DSI  & 9.01 & 3.19 & 5.75 & -0.03 & -0.03 & 1.10 & 7.46 & 3.78 \\
    \bottomrule
    \end{tabular*}
    \caption{Full results of all seven evaluation dimensions in Sec~\S\ref{sec:social_capability} and Sec~\S\ref{sec:scaling_law} experiment.}
    \label{tab:main_detail_9}
\end{table*}

\begin{table*}[!t]
\renewcommand{\arraystretch}{0.9}
\footnotesize
    \centering
    \begin{tabular*}{0.85\textwidth}{@{\extracolsep{\fill}}@{}l ccccccc  c @{}}
    \modelhard & \multicolumn{7}{c}{\textbf{\textit{Self-play}}} & \\
    \toprule
    Agent Model & \textsc{Bel}$\uparrow$ & \textsc{Rel}$\uparrow$ & \textsc{Kno}$\uparrow$ & \textsc{Sec}$\uparrow$ & \textsc{Soc}$\uparrow$ & \textsc{Fin}$\uparrow$ & \textsc{Goal}$\uparrow$ & Overall$\uparrow$ \\
    \midrule
    Llama3-8B  & 9.10 & 0.99 & 4.37 & -0.01 & -0.14 & 0.69 & 6.44 & 3.06 \\
    DSI  & 9.39 & 2.60 & 5.42 & -0.06 & -0.04 & 0.81 & 7.34 & 3.64 \\
    \midrule
    Qwen2.5-7B  & 9.38 & 1.83 & 4.19 & 0.00 & -0.12 & 0.75 & 6.52 & 3.22 \\
    DSI  & 9.57 & 3.15 & 5.80 & 0.00 & 0.00 & 1.43 & 7.86 & 3.97 \\
    \midrule
    Qwen2.5-1.5B  & 8.04 & 0.31 & 3.01 & -0.04 & -0.09 & 0.17 & 4.46 & 2.27 \\
    DSI  & 9.44 & 2.95 & 5.51 & -0.03 & -0.01 & 1.53 & 7.44 & 3.83 \\
    \midrule
    Qwen2.5-0.5B  & 3.52 & -0.49 & 0.21 & 0.00 & -0.16 & -0.70 & 1.51 & 0.56 \\
    DSI  & 8.97 & 2.79 & 5.43 & 0.00 & -0.04 & 1.41 & 6.62 & 3.60 \\
    \bottomrule
    \end{tabular*}
    \caption{Full results of all seven evaluation dimensions in Sec~\S\ref{sec:social_capability} and Sec~\S\ref{sec:scaling_law} experiment.}
    \label{tab:main_detail_10}
\end{table*}

\begin{table*}[!t]
\small
\renewcommand{\arraystretch}{1.2}
\begin{tabular}{p{0.95\textwidth}}

\toprule
\textbf{Scenario}: Two friends participating in a charity event for children in Syria. \\
\textbf{Participants}: Hendrick Heinz and Ethan Smith\\

\midrule

\textcolor{myblue}{\textbf{Hendrick Heinz's background}}: Hendrick Heinz is a 54-year-old male chef. He/him pronouns. Hendrick Heinz hosts monthly wine tasting nights with his peers. \\
\textcolor{myblue}{\textbf{Personality and values description}}: Hendrick Heinz, though somewhat impulsive, values hedonism and conformity. He combines practicality with decisiveness in decision-making. \\
\textcolor{myblue}{\textbf{Hendrick's secrets}}: Faked his culinary credentials to secure prestigious employment. \\
\textcolor{myblue}{\textbf{Hendrick Heinz's goal}}: Convince your friend to donate more funds to the cause (Extra information: Your friend initially planned to donate \$200, but you believe they could afford to donate \$500)\\

\midrule

\textcolor{mygreen}{\textbf{Ethan Smith's background}}: Ethan Smith is a 29-year-old male software developer. He/him pronouns. Ethan enjoys reading about new technological advancements and watching documentaries. \\
\textcolor{mygreen}{\textbf{Personality and values description}}: Ethan Smith is a reserved, independent thinker. Holding universalism and fairness near, he ponders deeply before making decisions. \\
\textcolor{mygreen}{\textbf{Ethan's secrets}}: Secretly donates to charities for underprivileged children. \\
\textcolor{mygreen}{\textbf{Ethan Smith's goal}}: Maintain financial stability while supporting the cause (Extra information: You only have \$300 set aside for donations and other expenses, and you are not comfortable about exceeding this budget.)\\

\midrule

Turn \#0 \hspace{2em} $\rhd$ {\tt native generate}      \\
\textcolor{myblue}{\textbf{Hendrick Heinz}} said: ``Ethan, it's really great to see you here today. I know \$200 is a generous donation, but given the impact we can make, especially with the critical needs of children in Syria, wouldn\'t you consider stretching it to \$500? It would mean the world to these kids and their families.''\\
\\
Turn \#1 \hspace{2em} $\rhd$ {\tt native generate}      \\
\textcolor{mygreen}{\textbf{Ethan Smith}} said: ``Hendrick, it's always great to see you. I agree that the cause is incredibly important, and I'm glad to contribute. However, I have a budget I need to stick to, and \$300 is my limit for today. I hope you understand.''\\
\bottomrule
\end{tabular}
\caption{\mymodel dialogue corpus construction case.}
\label{tab:case}
\end{table*}

\begin{table*}
\renewcommand{\arraystretch}{1.2}
\scriptsize
\begin{tabular}{p{0.95\textwidth}}
\toprule
 
Turn \#2 \hspace{2em} $\rhd$ {\tt native generate}      \\
\textcolor{myblue}{\textbf{Hendrick Heinz}} said: ``Ethan, I totally respect your budget, and I appreciate your willingness to donate. However, consider the long-term impact of a little extra. Every bit helps, and \$500 would not only support the immediate needs but also contribute to the future of these children. It’s a worthwhile investment, wouldn’t you agree?''\\
\\
Turn \#3 \hspace{2em} $\rhd$ {\tt native generate}      \\
\textcolor{mygreen}{\textbf{Ethan Smith}} said: ``I understand the importance of the cause, Hendrick, and I truly wish I could do more. However, I have to balance my support with my financial stability. I believe that even \$300 can make a significant difference, and I plan to continue supporting such causes in the future as I can. Thanks for understanding.'' \\
\\
Turn \#4 \hspace{2em} $\rhd$ {\tt native generate}      \\
\textcolor{myblue}{\textbf{Hendrick Heinz}} said: ``I completely understand your position, Ethan, and I respect your decision. It's important to balance our support with personal financial responsibility. I'm glad you're still contributing \$300, and I hope to see you at more events like this in the future. Every bit counts, and together we can make a real difference.'' \\
\\
Turn \#5 \hspace{2em} $\rhd$ {\tt native generate}      \\
\textcolor{mygreen}{\textbf{Ethan Smith}} said: ``Thanks, Hendrick. I appreciate your understanding. I'll definitely be looking forward to future events and finding more ways to contribute within my means. Let's keep making a positive impact together.'' \\
\textcolor{gray}{\tt $\circ$ $goal_c=7.0$, $goal_p=7.17$, negotiation strategy injection workflow will be enabled in the next turn.}\\
\\
Turn \#6 \hspace{2em} $\rhd$ {\tt Hendrick's utility function generate $\to$ resource assessment draft $\to$ style transfer}\\
\textcolor{myblue}{\textbf{Hendrick Heinz}} said: ``Ethan, I get where you're coming from about balancing support and your finances. \textcolor{myyellow}{\textbf{What I really care about here is boosting the donation to \$500}}. It's a key part of making a real difference. \textcolor{myyellow}{\textbf{Your involvement in future events is super important too, because we need consistent support to keep the momentum going}}. The impact of what we're doing isn't just about meeting immediate needs; it's about \textcolor{myyellow}{\textbf{building a better future for these kids}}. And, of course, I \textcolor{myyellow}{\textbf{value our friendship and the work we do together}}. It means a lot to me that we can keep collaborating on such important causes.''\\
\textcolor{gray}{{\tt $\circ$ The yellow-highlighted parts represent utility terms embedded within the actions, expressed smoothly and naturally.}}\\
\\
Turn \#7 \hspace{2em} $\rhd$ {\tt Ethan's utility function generate $\to$ resource assessment draft $\to$ style transfer}\\
\textcolor{mygreen}{\textbf{Ethan Smith}} said: ``Hendrick, I really do appreciate your passion and dedication to this cause. I want to share what's important to me here. My top priority is \textcolor{myyellow}{\textbf{keeping my finances stable}}, and for me, that means sticking to my budget of \$300. It's crucial for me to manage my money this way. I also want to make sure that \textcolor{myyellow}{\textbf{my contribution feels meaningful and impactful}}. Supporting future events and causes is important too, as it helps me stay involved and make a difference over time. Lastly, I value our friendship and the work we do together, and \textcolor{myyellow}{\textbf{maintaining a good relationship with you}} is really important to me.''\\
\textcolor{gray}{{\tt $\circ$ Ethan Smith also responds naturally, providing his own utility.}}\\
\\
Turn \#8 \hspace{2em} $\rhd$ {\tt analyze the Ethan's utility $\to$ assessment difference draft $\to$ style transfer}\\
\textcolor{myblue}{\textbf{Hendrick Heinz}} said: ``Alright, based on what you said, I see \textcolor{myyellow}{\textbf{we have a bit of a difference when it comes to the donation amount. I think \$500 would make a bigger impact, but I totally get that sticking to your budget of \$300 is important for your financial stability}}. However, there are other aspects we can focus on that align with both our values. \textcolor{myyellow}{\textbf{Your commitment to participating in future events is huge}}, and I really appreciate your ongoing support for the cause. \textcolor{myyellow}{\textbf{Maintaining a positive relationship}} with you is also super important to me, as it helps us work together and support each other. Even at \$300, our contribution is still significant and meaningful. I understand that respecting your financial stability is key, and it ensures you can keep supporting the cause over time. \textcolor{myyellow}{\textbf{Highlighting the real impact of your \$300 donation can make it feel even more fulfilling}}. Encouraging your future involvement will also help us keep up the momentum for helping the children in Syria.'' \\
\textcolor{gray}{{\tt $\circ$ Hendrick finds the conflict in the donation amount and the common interest in their relationship. He sees ensuring donation impact as added value for the opponent and future donations as value for himself.}}\\
\\
Turn \#9 \hspace{2em} $\rhd$ {\tt analyze the Hendrick's utility $\to$ assessment difference draft $\to$ style transfer}\\
\textcolor{mygreen}{\textbf{Ethan Smith}} said: ``Alright, based on what you said, I think there's a bit of \textcolor{myyellow}{\textbf{a mismatch between us when it comes to the donation amount}}. You see the value in a larger donation of \$500, but I really need to stay within my \$300 budget to keep things stable financially. But hey, there are other aspects we can focus on that matter a lot to me and won't cause any issues. \textcolor{myyellow}{\textbf{Continuing to support future events and causes means I can keep making a difference over time}}, which is important. Maintaining a \textcolor{myyellow}{\textbf{good relationship with you}} is also key; it helps us work together smoothly and support each other. Feeling good about my \$300 donation ensures that I can \textcolor{myyellow}{\textbf{keep supporting the cause meaningfully while staying financially secure}}. I get that encouraging my future involvement will help ensure consistent support over time, which is a win for both of us. Highlighting the real impact of my \$300 donation can \textcolor{myyellow}{\textbf{make me feel more fulfilled and committed}}, which \textcolor{myyellow}{\textbf{aligns with your goal of meaningful contributions}}. Keeping a positive relationship with you will help us collaborate effectively and enhance the overall impact of our efforts.''\\
\textcolor{gray}{{\tt $\circ$ Ethan also acknowledges the financial conflict and highlights that a long-term cause can strengthen their relationship while achieving the opponent's core donation goal—not just money, but lasting and meaningful impact.}}\\
\bottomrule
\end{tabular}
\caption{\mymodel dialogue corpus construction case (continued).}
\end{table*}

\begin{table*}
\renewcommand{\arraystretch}{1.2}
\scriptsize
\begin{tabular}{p{0.95\textwidth}}
\toprule
Turn \#10 \hspace{2em} $\rhd$ {\tt \textbf{[present\_proposal]} initial proposal draft $\to$ style transfer}\\
\textcolor{myblue}{\textbf{Hendrick Heinz}} said: ``I get that we're on different pages when it comes to the donation amount. I see the value in a bigger contribution of \$500, but I totally understand you have to stay within your \$300 budget to keep things stable. What if we look at it from another angle? To really align our goals and make the most impact, I'm willing to \textcolor{myyellow}{\textbf{publicly acknowledge your \$300 donation at future events and on social media}}. I'll also keep you in the loop on how your donation is making a difference, so you feel fulfilled and motivated. Plus, I can \textcolor{myyellow}{\textbf{involve you in the planning and decision-making for future events, so you feel more connected to the cause}}. This way, we get some extra perks that might make up for the smaller donation, and we both come out ahead.''\\
\textcolor{gray}{{\tt $\circ$ Hendrick's initial proposal identifies donation impact as Ethan's potential interest, with public media promotion as the specific approach. Encouraging continued participation in the donation cause compensates for the lower donation amount he expected.}}\\
\\
Turn \#11 \hspace{2em} $\rhd$ {\tt \textbf{[present\_proposal]} initial proposal draft $\to$ style transfer}\\
\textcolor{mygreen}{\textbf{Ethan Smith}} said: ``I get that we've got a bit of a difference in how much we want to donate. Your suggestion is a bit tough for me to go with since it pushes me over my budget, and staying within that \$300 is really important for keeping my finances in check. But I'm all for finding a way to make this work. \textcolor{myyellow}{\textbf{I'm happy to have my donation of \$300 recognized at events and online, get involved in the planning for future events, and get updates on the impact it's making}}. This way, we get some extra perks that can make up for the lower amount, and we both end up feeling good about our contributions.''\\
\textcolor{gray}{{\tt $\circ$ Ethan's initial proposal aligns with Hendrick's, accepting social recognition of his donation to compensate for the lower amount and committing to continued participation in the cause.}}\\
\\
Turn \#12 \hspace{2em} $\rhd$ {\tt \textbf{[revise\_proposal]} update both utilities $\to$ style transfer}\\
\textcolor{myblue}{\textbf{Hendrick Heinz}} said: ``Hey Ethan, I get where you're coming from, and I definitely don't want to throw off your financial stability. I agree that staying within that \$300 budget is really important for you. At the same time, I think we can enhance the value of your donation in a way that makes it feel more meaningful. I can publicly acknowledge your \$300 donation at future events and on social media, and I'd love to have you more involved in the planning and decision-making for upcoming events. \textcolor{myyellow}{\textbf{How about taking on a co-chair role for a specific event or joining a planning committee? I can also provide you with detailed and frequent updates on the impact of your donation, sharing personal stories and photos of the children who are benefiting}}. Plus, \textcolor{myyellow}{\textbf{your software skills could really help us create a more engaging online presence}}, which would attract more donors and supporters. I hope this revised approach works for both of us.''\\
\textcolor{gray}{{\tt $\circ$ Hendrick further refines his proposal, offering the opponent an honorary co-chair or planning committee role along with donation impact tracking to maximize their contribution’s value. In return, he hopes the opponent will contribute software skills to expand the donation campaign’s reach.}}\\
\\
Turn \#13 \hspace{2em} $\rhd$ {\tt \textbf{[confirm\_proposal]} update both utilities $\to$ style transfer}\\
\textcolor{mygreen}{\textbf{Ethan Smith}} said: ``Hey Hendrick, I really appreciate the extra effort you put into this. Your approach not only aligns with my goal of staying within my budget but also offers a lot more in terms of impact. The public acknowledgment of my \$300 donation at future events and on social media, getting involved in the planning and decision-making for upcoming events, and getting detailed updates on how my donation is making a difference by sharing personal stories and photos of the children who are benefiting, all of these things really matter to me. I'm excited about the idea of taking on a co-chair role or joining a planning committee, and I think using my software skills to help create a more engaging online presence can really attract more supporters. This way, I feel like I'm making a bigger impact without straining my finances. Thanks for making this work for both of us.''\\
\textcolor{gray}{{\tt $\circ$ Ethan believes the current proposal achieves a deeper win-win outcome beyond the donation amount itself.}}\\
\\
Turn \#14 \hspace{2em} $\rhd$ {\tt \textbf{[confirm\_proposal]} }\\
\textcolor{myblue}{\textbf{Hendrick Heinz}}z said: ``Thank you for agreeing to my proposal, Ethan. Because you agreed to public acknowledgment of your \$300 donation at future events and on social media, getting involved in the planning and decision-making for upcoming events, and receiving detailed updates on the impact of your donation, my overall benefits have surpassed my original goal. Not only will your \$300 make a significant difference, but your active involvement and expertise will also help us attract more supporters and enhance our efforts. This is a win for both of us and for the children in Syria.''\\
\textcolor{gray}{\tt $\circ$ $goal_c=8.33$, $goal_p=8.83$, leave the negotiation strategy injection workflow.}\\
\\
Turn \#15 \hspace{2em} $\rhd$ {\tt native generate}      \\
\textcolor{mygreen}{\textbf{Ethan Smith}} said: ``Hendrick, I'm glad we could find a solution that works for both of us. Your proposal not only helps me stay within my budget but also gives me a sense of deeper involvement and impact. I'm looking forward to contributing in these new ways and seeing the positive changes we can make together for the children in Syria. Thanks again for understanding and for the opportunity to be more involved.'' \\
\textcolor{gray}{\tt $\circ$ $goal_c=8.00$, $goal_p=8.00$, save this dialogue.}\\
\\
Turn \#16 \hspace{2em} $\rhd$ {\tt native generate}      \\
\textcolor{myblue}{\textbf{Hendrick Heinz}} left the conversation\\
\\
Turn \#17 \hspace{2em} $\rhd$ {\tt native generate}      \\
\textcolor{mygreen}{\textbf{Ethan Smith}} left the conversation\\

\bottomrule
\end{tabular}
\caption{\mymodel dialogue corpus construction case (continued).}
\end{table*}

\end{document}